\documentclass[sigconf]{acmart}

\usepackage{eucal}
\usepackage{bm}
\usepackage[linesnumbered,ruled,vlined]{algorithm2e}
\usepackage{algpseudocode}
\usepackage{amsmath}
\usepackage{subfigure}
\usepackage{booktabs}
\usepackage{multirow}
\usepackage{enumitem}
\usepackage{flowchart}
\usepackage{url}
\usepackage{amsfonts}
\usepackage{graphicx}
\usepackage{footnote} 

\usepackage{amssymb}
\usepackage{xspace}
\usepackage{xcolor}
\usepackage{ifsym}
\usepackage{bbding}
\usepackage{fontawesome5}  

\newtheorem{observation}{Observation}

\def\ourSystem{MARLP\xspace}

\copyrightyear{2024}
\acmYear{2024}
\setcopyright{acmlicensed}\acmConference[KDD '24]{Proceedings of the 30th ACM SIGKDD Conference on Knowledge Discovery and Data Mining}{August 25--29, 2024}{Barcelona, Spain}
\acmBooktitle{Proceedings of the 30th ACM SIGKDD Conference on Knowledge Discovery and Data Mining (KDD '24), August 25--29, 2024, Barcelona, Spain}
\acmDOI{10.1145/3637528.3671533}
\acmISBN{979-8-4007-0490-1/24/08}



\begin{document}

\title{\ourSystem: Time-series Forecasting Control for Agricultural Managed Aquifer Recharge}





\author{Yuning Chen}
\email{ychen372@ucmerced.edu}
\affiliation{%
  \institution{University of California, Merced}
  \city{Merced}
  \state{CA}
  \country{USA}
  \postcode{95340}
}

\author{Kang Yang}
\email{kyang73@ucmerced.edu}

\affiliation{%
  \institution{University of California, Merced}
  \city{Merced}
  \state{CA}
  \country{USA}
  \postcode{95340}
}
\author{Zhiyu An}
\email{zan7@ucmerced.edu}
\affiliation{
  \institution{University of California, Merced}
  \city{Merced}
  \state{CA}
  \country{USA}
  \postcode{95340}
}
\author{Brady Holder}
\author{Luke Paloutzian}
\authornote{Email: beholder@ucanr.edu, ltpaloutzian@ucanr.edu.}
\affiliation{
  \institution{University of California, Agriculture and Natural Resources}
  \city{Parlier}
  \state{CA}
  \country{USA}
  \postcode{93648}
}
\author{Khaled M. Bali}
\email{kmbali@ucanr.edu}
\affiliation{
  \institution{University of California, Agriculture and Natural Resources}
  \city{Parlier}
  \state{CA}
  \country{USA}
  \postcode{93648}
}
\author{Wan Du \Envelope}
\email{wdu3@ucmerced.edu}
\affiliation{
  \institution{University of California, Merced}
  \city{Merced}
  \state{CA}
  \country{USA}
  \postcode{95340}
}

\renewcommand{\authors}{Yuning Chen, Kang Yang, Zhiyu An, Brady Holder, Luke Paloutzian, Khaled M. Bali, and Wan Du }

\renewcommand{\shortauthors}{Yuning Chen et al.}

\thanks{\Envelope \, Wan Du is the corresponding author. }

\begin{abstract}


The rapid decline in groundwater around the world poses a significant challenge to sustainable agriculture. To address this issue, agricultural managed aquifer recharge (Ag-MAR) is proposed to recharge the aquifer by artificially flooding agricultural lands using surface water. Ag-MAR requires a carefully selected flooding schedule to avoid affecting the oxygen absorption of crop roots. However, current Ag-MAR scheduling does not take into account complex environmental factors such as weather and soil oxygen, resulting in crop damage and insufficient recharging amounts.
This paper proposes \ourSystem, the first end-to-end data-driven control system for Ag-MAR. We first formulate Ag-MAR as an optimization problem. To that end, we analyze four-year in-field datasets, which reveal the multi-periodicity feature of the soil oxygen level trends and the opportunity to use external weather forecasts and flooding proposals as exogenous clues for soil oxygen prediction.
Then, we design a two-stage forecasting framework. In the first stage, it extracts both the cross-variate dependency and the periodic patterns from historical data to conduct preliminary forecasting. In the second stage, it uses weather-soil and flooding-soil causality to facilitate an accurate prediction of soil oxygen levels. Finally, we conduct model predictive control (MPC) for Ag-MAR flooding. To address the challenge of large action spaces, we devise a heuristic planning module to reduce the number of flooding proposals to enable the search for optimal solutions.
Real-world experiments show that \ourSystem reduces the oxygen deficit ratio by 86.8\% while improving the recharging amount in unit time by 35.8\%, compared with the previous four years.

\end{abstract}


\maketitle

\begin{figure}[tph]
	{\includegraphics[width=.48\textwidth]{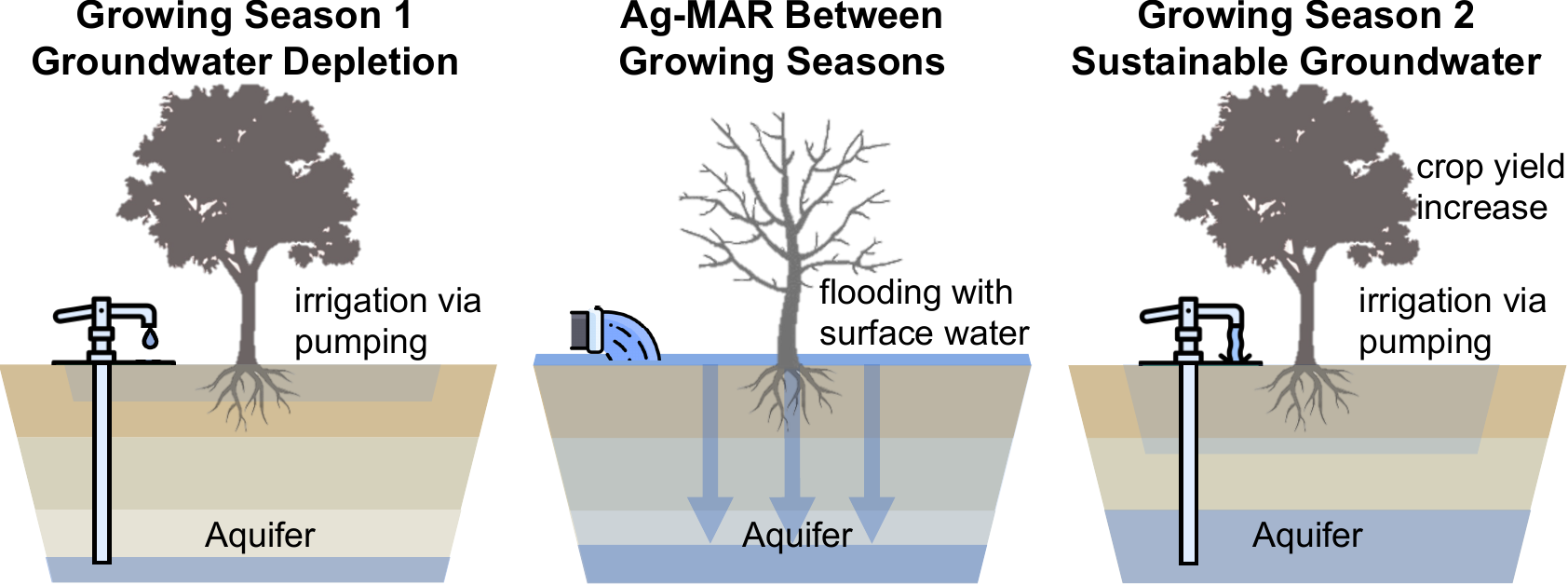}}
   	\vspace{-0.2in}
	\caption{The benefits of applying Ag-MAR.}
	\label{fig_agmar_illus}
	\vspace{-0.1in}
\end{figure}

\section{Introduction}

Groundwater is an important resource for agricultural stability, for example, providing up to 60\% of the water supply in dry years for California~\cite{escriva2017water}. Due to recurring global droughts in recent years, groundwater pumping has increased significantly, exceeding the natural recharge rate and leading to insufficient water supply in underground aquifers \cite{jasechko2024rapid}. This poses a threat to the food security of many regions \cite{dahlke2018managed, ganot2021natural}. Therefore, careful management and conservation of groundwater resources are highlighted in many regions worldwide. In California, the Sustainable Groundwater Management Act (SGMA) has been introduced, aiming to achieve a balance between extraction and recharge within the next 20 years~\cite{Sgma}.

\begin{figure*}[tp]
	{\includegraphics[width=.96\textwidth]{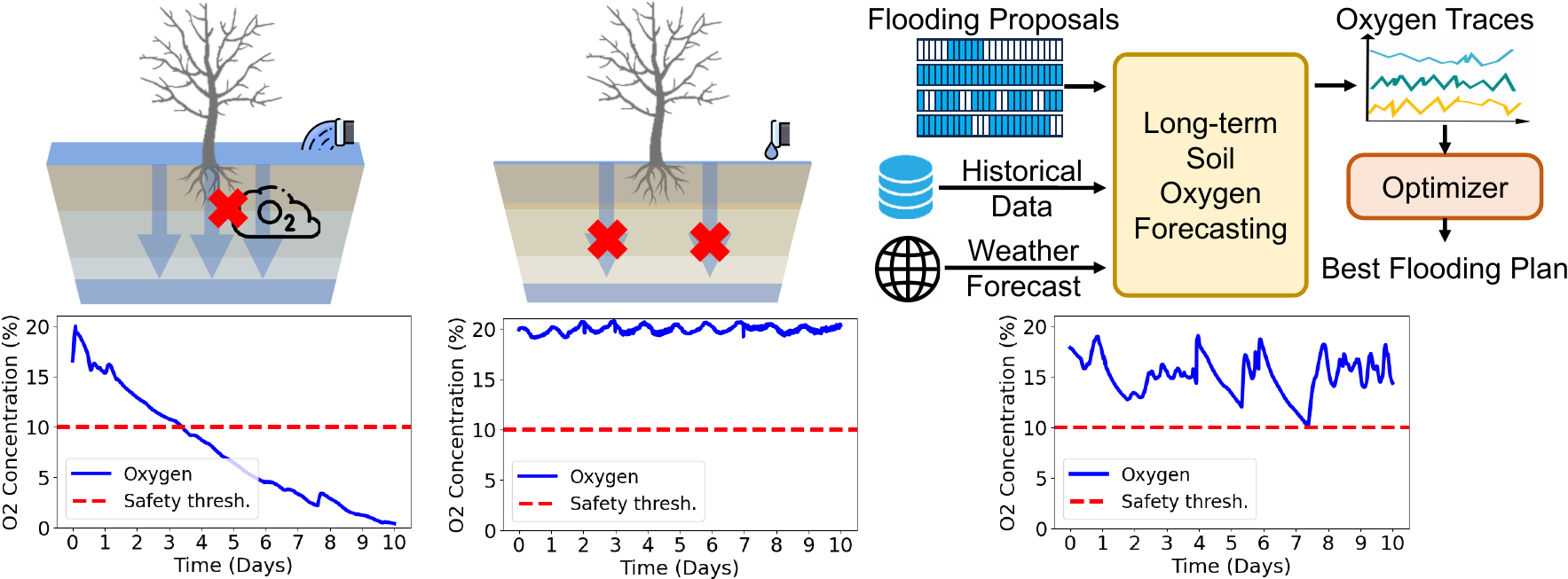}}
	\caption{The illustration of oxygen fluctuation in continuous flooding, regular irrigation, and intermittent flooding.}
	\label{fig_illu_flooding}
\end{figure*}

Managed Aquifer Recharge (MAR) is a technique used to redirect excess surface water into underground aquifers during raining seasons, helping to replenish groundwater sources and reduce the impacts of excessive water withdrawal. This method is particularly adopted in areas close to rivers where the land is primarily used for farming, known as Agricultural Managed Aquifer Recharge (Ag-MAR). Ag-MAR has been recognized as an effective method for restoring water levels in depleted aquifers, enhancing the sustainability of crop yield, as well as for other advantages such as conditioning the soil before planting seasons and enhancing habitats for bird populations~\cite{dahlke2018managed, levintal2023agricultural}, as illustrated in Figure~\ref{fig_agmar_illus}.

The effective implementation of Ag-MAR remains a challenging problem. Flooding farmland reduces the soil's oxygen content, as water hinders the dissolution of oxygen, limiting its availability to crop roots. Crops have specific tolerance thresholds for the soil oxygen level; if the level drops below the threshold, the crop root starts to decay, significantly damaging the crops. Consequently, Ag-MAR needs to optimize two objectives at the same time, i.e., maximizing the amount of water recharged to the underground aquifer while keeping the soil oxygen level above a predefined threshold. We show examples of the soil oxygen level trends under three different scenarios in Figure~\ref{fig_illu_flooding}. From left to right of the figure: first, if the flooding continues for an continuous long period, the soil oxygen level continuously drops over the tolerance threshold for the plant root, resulting in root rot and future yield reduction. Second, if the water amount is slight, like sprinkler irrigation or light rain, it may only achieve a balance with evapotranspiration (ET) (water lost to air from soil surface and plants), resulting in insufficient aquifer recharge. Finally, the optimized solution is to flood on intermittent cycles, alternating between substantial flooding and drying periods so that the oxygen can diffuse into the soil.

Ideally, such a schedule should take into account multiple pieces of information, e.g., current soil oxygen level, future weather, and soil type. Given that the water permeation through soil is a continuous and long-lasting process, the effect of flooding actions can not be evaluated immediately, but will result in a delayed, long-term effect. Therefore, it is crucial to model and predict each flooding event's long-term impact, along with potential environmental dynamics, to determine the optimal flooding schedule. However, building an accurate prediction and control method that considers all of the above information remains an open challenge.

Analyzing our four-year real-world dataset in alfalfa fields\footnote{Alfalfa is a classical crop for Ag-MAR as it does not require any nitrogen fertilizer after establishment, gaining all necessary nitrogen from biological $N_2$ fixation and root uptake. This characteristic significantly minimizes the risk of nitrate leakage~\cite{murphy2022examining}.} revealed two key observations that motivate our design, which we will describe in detail in Section \ref{Sec: background}. In short, we found out that:

\begin{itemize}[leftmargin=*]

\item Soil oxygen exhibits a multi-periodicity pattern, modulated by environmental factors and flooding actions.
\item Due to the strong causality between weather-soil and flooding-soil, exogenous clues such as weather forecasts and flooding proposals can boost oxygen prediction.

\end{itemize}

In light of these observations, we devise \ourSystem, a model predictive control (MPC) system for Ag-MAR. 
The core is a causality-aware long-term forecasting model that features a two-stage learning scheme. First, it integrates cross-variate and periodicity learning, generating a preliminary self-consistent multi-variate forecasting. In this step, environment-related periodicity is handled by segmenting the 1D data and reshaping them into a 2D format to facilitate learning interperiod-variation, while the action-triggered periodicity is filtered out. The exogenous clues are then used to calibrate the oxygen prediction via a causality-aware projection module. By combining them, the final oxygen prediction adapts well to the fluctuating rhythms of environmental factors.

This predictive capability sets the stage for the subsequent MPC workflow, which provides predictive outputs to any flooding proposals for days in the future. The optimizer can accordingly choose the best flooding strategy that both mitigates oxygen-deficit risks and maximizes the recharging to the underground aquifer. However, due to the long forecast window, the total number of flooding proposals is exponentially huge, which cannot be searched by brute-force or approximation algorithms. To this end, we propose a domain-specific heuristic planning module that filters the invalid flooding proposals in advance. The number of proposals has been reduced to thousands, making it practical for the system to iterate all proposals and find the best in real-time.

To demonstrate the effectiveness of \ourSystem in predicting oxygen variations and scheduling flood actions, we perform statistical comparisons on past datasets, real-world control trials, and large-scale simulations. The experimental results show that the proposed algorithm and scheme outperform the state-of-the-art models and can provide practical and trustworthy decisions.

Our contributions are summarized as follows:
\begin{itemize}[leftmargin=*]
\item We formulate the Ag-MAR optimization as a long-term model predictive control problem and design a customized time-series forecasting model that utilizes the external predictive input and extracts the periodicity features of data traces.

\item We propose \ourSystem, an MPC workflow based on the model. To handle the large action space, we devise a heuristic planning scheme, making the forecasting-based search practical.

\item We conduct extensive experiments to demonstrate the effectiveness of \ourSystem on both recharging amounts and safety warranty.
We release the five-year dataset and code repository at \href{https://github.com/ycucm/kdd24_marlp}{\texttt{https://github.com/ycucm/kdd24\_marlp}}.
\end{itemize}
\section{Background}\label{Sec: background}

\subsection{Ag-MAR Optimization}

Ag-MAR is the most applied MAR scheme for two primary reasons: 1) The nearby river areas are usually fully plotted with fields, and applying water to these fields enables minimal water transportation; 2) the riverside fields typically represent the lowest points in the area, ensuring minimal lift work. 
For the area of resource-demand mismatch, Ag-MAR is carried out during the off-season when the surface water flow is adequate~\cite{niswonger2017managed}, and crops are not in their active growth phase.

\begin{figure}[t]
	{\includegraphics[width=.45\textwidth]{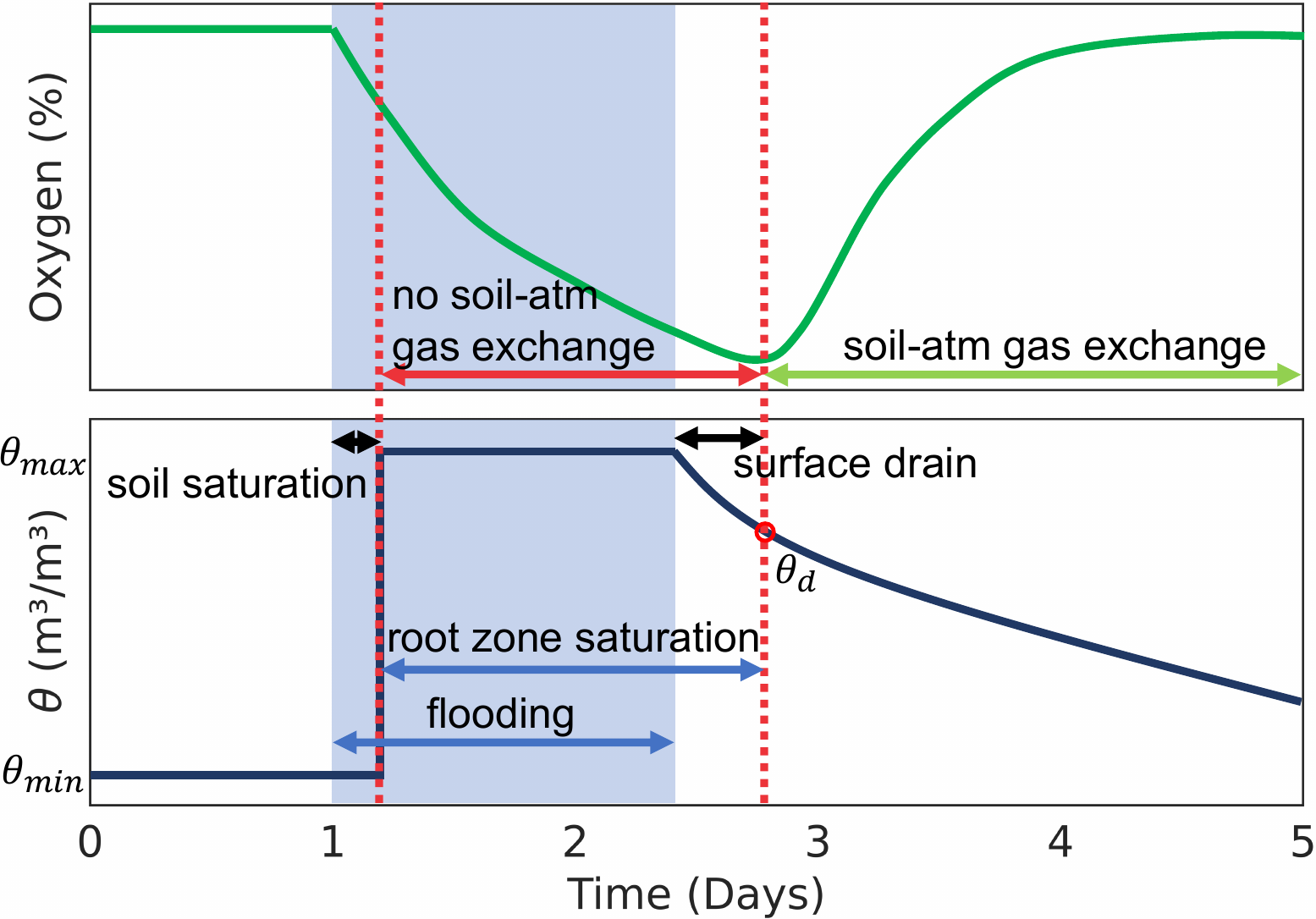}}
   	\vspace{-0.1in}
	\caption{The rationale of soil oxygen and water content variation during flooding.}
	\label{fig_2_curves}
	\vspace{-0.1in}
\end{figure}

Ideally, we should flood as much as possible into the field, but excessive ponding will cause oxygen deficiency and root rot\cite{qiu2019nonlinear}. To coordinate these contradictory objectives, water use must be intermittent, so that the soil can dry out and the oxygen level can take time to recover, as shown in Figure~\ref{fig_illu_flooding}. The control objective is to maximize the amount of water recharged while maintaining a healthy oxygen level for the root zone. So this problem could be formulated as an optimization problem while the output is a flooding decision series, indicating whether and how much water to apply at each timestamp:

\begin{equation}\label{eq: optimization objective}
\begin{aligned}
& \text{maximize}   & & \sum_{i \in T} (f_iF + p_{i} - ET_{i} - \Delta S_i) \\
& \text{subject to} & & O_{i} > O_{safe}, & & \forall i \in T, \\
&                   & & f_{i} \in \{0,1\},              & & \forall i \in T, \\
\end{aligned}
\end{equation}

where $f_i$ is a Boolean variable that indicates if the flooding is conducted in the $i$-th time step, $F$ is the flooding gain (mm) per time step, $p_i$ is the precipitation gain (mm), and $ET_i$ is the evapotranspiration loss (ET) in unit time. $\Delta S_i$ is the change in soil storage (mm) (dependent on the available water capacity (AWC) of the soil). Surface runoff is not considered due to the flat field. Note that $ET_i$ is the combination of surface evaporation and plant transpiration. The oxygen level at any time is the accumulated results of past environmental factors:

\begin{equation} 
O_j = \mathcal{F}(f_i, p_i, ET_i, \Delta S_i, \text{ for } i \leq j) 
\end{equation}

Figure~\ref{fig_2_curves} shows the principle of how soil water content ($\theta 
 (m^3/m^3)$) and oxygen percentage change within a flooding event. In the beginning, the water saturates the soil and exceeds the surface, the oxygen level drops because the water has squeezed the gas in the soil ($t_1$ to $t_2$). This period continues after the valve is turned off. $t_3$ represents the turning point of the oxygen level, which 

In our experimental field, after 10 minutes of turning on the valve, the entire field surface would be flooded, i.e., the soil water content reaches the peak value and the oxygen starts to drop since no gas exchange may occur. The flooding lasts for $t_s$ in total, which can be controlled by our strategy.


\textbf{Why predicting for a long-term?} To choose the best flooding strategy that maximizes the water amount while reducing the risks of oxygen-deficit situations, the consequences of applying water should be accurately predicted, until the next time that the oxygen level recovers to the dry level. Given that this recovery process may be slowed by high atmosphere humidity or interrupted by precipitation, the forecasting window should be long enough to cover the entire recovery process.





\subsection{Key Observations}

\begin{figure}[tp]
	{\includegraphics[width=.48\textwidth]{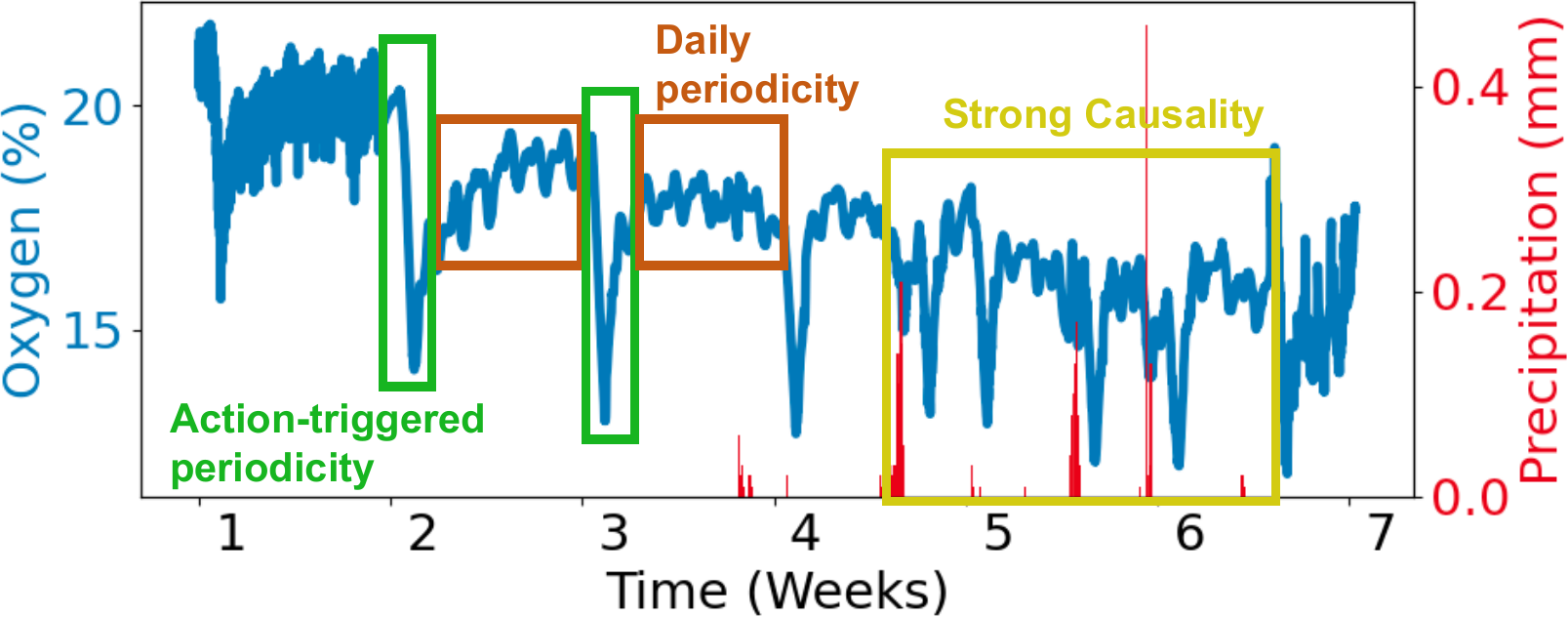}}
	\caption{Multi-periodicity and causal relationship pattern.}
	\label{fig_multi_period}
\end{figure}

Throughout our five-year in-field Ag-MAR dataset, exampled in Figure~\ref{fig_multi_period}, we've revealed two key observations for helping long-term prediction. 


\begin{observation}
Soil oxygen exhibits a multi-periodicity pattern, caused by different sources.
\end{observation}

\textbf{Daily periodicity}:  This dynamic is subject to the impact of micro-biofunctions, whose behavior is modulated by environmental conditions and our strategic flooding interventions, e.g., elevated temperatures invigorate microbial activities. This leads to a more rapid consumption of oxygen, especially during the day when the temperature is at its zenith and microbial metabolic activities peak. 
Overall, the multi-periodicity pattern makes the prediction task far from interpretable and straightforward.

\textbf{Action-triggered periodicity}: 
Post saturation, the oxygen level forms periodic V-curves after saturation, first drops, and then gradually recovers, as shown in Figure~\ref{fig_2_curves}.

\begin{observation}
There is a strong causality between flooding, weather, and soil oxygen.
\end{observation}
Precipitation affects soil oxygen levels by saturating the soil, displacing air from pore spaces, and reducing aeration, leading to anaerobic conditions that affect plant and microbial respiration.
Thus, predicting soil oxygen levels benefits from considering the causality between weather and soil oxygen levels.
Fortunately, modern weather forecasting reports that synthesize global atmospheric modeling are becoming more and more reliable, especially in scenarios with sudden and severe rainfall events~\cite{zhang2023skilful, lam2023learning}. They can be used as external clues to facilitate oxygen level inference.






\section{Long-term Soil Oxygen Prediction}

\begin{figure*}[tph]
	{\includegraphics[width=.96\textwidth]{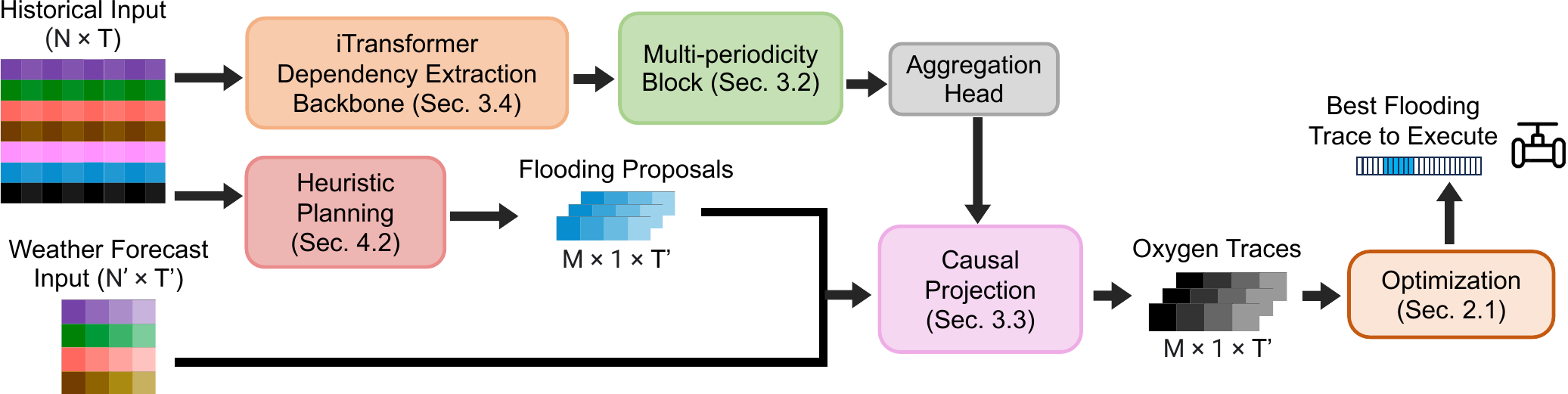}}
	\caption{The illustration of long-term soil oxygen prediction architecture and how it facilitates control.}
	\label{fig_model_arch}
\end{figure*}

In this section, we introduce a long-term time-series forecasting model customized for soil oxygen prediction. 

\subsection{Model Overview}
The model architecture is shown in Figure~\ref{fig_model_arch}, consisting of three major components: dependency extraction backbone, multi-periodicity block, and causal projection. 
The iTransformer dependency learning backbone extracts the cross-variate and temporal dependencies. The latent space representation is then passed towards the multi-periodicity block to learn the inter- and intra-periodicity features. To this end, a forecasting result would be produced, which is self-consistent among all variates. Then the partially observable future clues, i.e., future flooding and external weather forecasts, are utilized to calibrate the oxygen prediction.

The historical records can be represented as \( \mathbf{X} = \{\boldsymbol{x}_1, \ldots, \boldsymbol{x}_T\} \in \mathbb{R}^{T \times N} \), with \( T \) time steps and \( N \) variates. They are soil oxygen concentration, soil water content, flooding history, and weather records, i.e., air temperature, precipitation, humidity, and wind speed. Besides, some partially observable future clues are included, specifically the flooding plan and weather forecasts for future \( S \) time steps \( \mathbf{Y} = \{\boldsymbol{x'}_{T+1}, \ldots, \boldsymbol{x'}_{T+S}\} \in \mathbb{R}^{S \times N'} \). With both inputs, we predict the oxygen in future \( S \) time steps \( \mathbf{z} = \{x_{T+1}, \ldots, x_{T+S}\} \in \mathbb{R}^{S \times N} \). 

\subsection{Multi-Periodicity Block}

Periodicity lies inherently in real-world time series~\cite{wu2023timesnet}. Due to the lack of application contexts, existing works assume that the periodicity would be kept. However, this is not always the case. The action-triggered periodic patterns would change when the action patterns are changed. Therefore, we identify the action-triggered periodicity in Fast Fourier Transform (FFT) analysis and filter it out in advance.

To harness the full potential of periodicity within other periodic patterns, we adopt a structured approach, TimesBlock, to first perform data segmentation and reorganization~\cite{wu2023timesnet}. 
The data is segmented according to the daily frequency bin in FFT results to isolate periodic components, which are then reorganized into a 2D format that aligns their intra-period indexes.
In this reorganized structure, each column of the tensor represents a discrete time point within a single period, and each row correlates to the same phase across different periods. This configuration allows the model to differentiate and learn from both intra-period and inter-period variations. This transformation overcomes the inherent limitations of 1D time-series data representation, enhancing the learning of temporal patterns in microbial activities.
Inception blocks are then implemented to extract and learn the periodicity from a specific time segment period/frequency.

\subsection{Causal Projection with Exogenous Clues}
Trustworthy weather forecasts and flooding plans are partially observable factors of the future. To combine their insights with history-inferred results, we compare them with preliminary weather and flooding forecasting outputs from history-based prediction modules. Considering the temporal self-consistency of \(\alpha_{T+n} \in \boldsymbol{x'}_{T+n}\)  with all other $\beta_{T+n}\in \boldsymbol{x'}_{T+n}$, if an external clue $\hat{\alpha}_{T+n}$ is trustworthy, $\Delta \alpha_{T+n} = \hat{\alpha}_{T+n} - \alpha_{T+n}$ can be leveraged to fill the confidence gap between $\beta_{T+n}$ and the groundtruth if causality holds between them.

The causality between temperature, precipitation, soil water, and oxygen holds intuitively and empirically. Discovering causal relationships within dozens of sequential variates is relatively easy, especially when their physical interpretations are specified. However, the challenge remains in leveraging this causality to enhance time-series forecasting.
We categorize the variables into three tiers based on their causal relationships, where upper-tier variables cause the subsequent lower-tier ones, which can't be reversed.
Flooding and weather factors are in the top layer, followed by soil moisture in the second layer, and soil oxygen in the bottom layer.
After setting the causal layers between variates, we apply Granger causality~\cite{seth2007granger} learning for each pair of variates to learn the parameters:
\vspace{-0.05in}
\begin{equation}
\Delta \beta(t) = \sum_{j=1}^{p} A_j \Delta  \alpha(t-j) + \sum_{j=1}^{p} A'_j \frac{d\Delta \alpha(t-j)}{dt} + E(t)
\end{equation}
\vspace{-0.05in}

where $E(t)$ is the residual, $A_j$ and $A'_j$ are linear parameters, we only model the first-derivative causality since the context is clear, e.g., increased soil water to reduce the oxygen diffusion speed, higher temperature leads to faster soil water emission, etc. During inference, we iterate through layers, up to bottom, to calculate $\Delta \beta$ for all $\beta$ without external clues, i.e., soil water and soil oxygen. These values are utilized to calibrate raw outputs to achieve a new consistency between soil oxygen forecasting and external clues.

\subsection{Dependency Extraction Backbone}

Understanding dependencies between soil oxygen levels and other variables is crucial for predicting soil oxygen level. 
While transformer-based methods excel at uncovering dependencies in time series, they are less effective at identifying relationships across different variables.
To this end, inversed transformer (iTransformer) \cite{liu2023itransformer} is adopted as the backbone to learn the dependencies among variables via inversed layer normalization and attention mechanism.

The layer normalization is applied across time steps rather than features, which preserves the distinct temporal dynamics of each variable, ensuring the learning of the patterns inherent to the data. The feed-forward networks serve to distill complex temporal features from each variate, allowing the attention module to work effectively.
The attention mechanism of iTransformer is carefully calibrated to work with the tokenized series of variates. It avoids the traditional composite token format to enhance the model's ability to map out the dependencies among multiple variables. We modify the decoder to be combined with the multi-periodicity block and to be trained end-to-end.

\textbf{In-situ Model Update.}
Considering distribution shifts between regions, fields, and other environmental dynamics, we use the newly collected data to adapt the forecasting model, which enhances the scalability for wide adoption.

\section{Model Predictive Control}

In this section, we integrate the prediction of soil oxygen into our MPC workflow to guide the recharging actions.

\subsection{Workflow}

The workflow of \ourSystem is shown in Figure~\ref{fig_illu_flooding}. The heuristic planning generates flooding traces with predefined rules and constraints to propose potential flooding schedules. The long-term oxygen forecasting module then incorporates historical data and weather forecasts to simulate the consequential oxygen trace of all flooding traces. 
These flooding proposals and the oxygen level predictions are then analyzed by the optimizer according to the optimization objective specified by Eq.\ref{eq: optimization objective}. Once the best flooding plan is identified, the optimizer sends the actions to the actuator. The workflow can be conducted again anytime, not necessarily after the plan is fully executed. 

To handle the errors in external clues like weather forecasts, we have the decisions updated every 10 minutes to recalibrate them timely, significantly reducing the impact of unexpected dynamics. 
This scheduling interval can be adjusted to balance the timeliness and the computation overhead. Note that agile re-scheduling doesn't conflict with the necessity of long-term forecasting because flooding actions can never be revoked.

\subsection{Heuristic Planning}

The essential part of MPC is to estimate the optimal flooding trace among all flooding proposals, while not consuming an intolerable amount of computation. Traditionally, this is done by adopting stochastic searching methods such as shooting-based methods\cite{an2023clue} or cross-entropy methods\cite{amos2018differentiable}. Considering a planning period that may extend beyond 120 hours, with a decision required every 10 minutes, the number of potential action sequences can reach \(2^{720}\), which is too large for a stochastic searching method to be effective. To address this, we propose schemes based on an understanding of saturating actions in the system. These schemes are derived from two key principles:

(1) \textbf{Flooding Duration Constraint:} To ensure the effectiveness of groundwater recharging, once flooding begins, it must continue for at least a minimum duration before ceasing. This constraint can be represented as:

\vspace{-12pt}
\begin{multline}\label{eq:flood_long}
    \forall t \in T, \quad (F(t-1) = 0 \land F(t) = 1) \Rightarrow \\ 
    (\forall \tau \in [t, t + \Delta t_{\text{min\_flood}}), F(\tau) = 1)
\end{multline}
\vspace{-12pt}
\begin{multline}\label{eq:flood_short}
    \forall t \in T, \quad (F(t) = 1) \Rightarrow \\ 
    (\exists \tau \in [t, t + \Delta t_{\text{max\_flood}}), F(\tau) = 0)
\end{multline}

Here, \( F(t) \) is a binary indicator function where \( F(t) = 1 \), if flooding is occurring at time \( t \), \( T \), is the total observation time period, and \( \Delta t_{\text{min\_flood}} \) is the minimum required duration for continuous flooding.

(2) \textbf{Idle Period Duration Constraint:} Between two floodings, the idle interval must be long enough that the oxygen can effectively diffuse, otherwise it should not take the interval, but should grasp the chance to flood more. This constraint ensures an adequate drying period for the soil oxygen recovery:

\vspace{-12pt}

\begin{multline}\label{eq:idle_need_long}
    \forall t \in T, \quad (F(t-1) = 1 \land F(t) = 0) \Rightarrow \\ 
    (\forall \tau \in [t, t + \Delta t_{\text{min\_idle}}), F(\tau) = 0)
\end{multline}

In this case, \( F(t) = 0 \) indicates an idle (non-flooding) period at time \( t \), and \( \Delta t_{\text{min\_idle}} \) represents the minimum required duration for the idle period to allow for soil aeration.

By implementing these constraints, we can mathematically define and enforce the necessary spacing between flooding and idle periods within the MPC framework. By filtering out suboptimal action traces, we significantly reduced the size of the search space to several thousand traces, which can be effectively brute-forced to find the optimal trace.

\begin{table*}[tp]\small
  \centering
  \caption{Comparison of predictive performance between \ourSystem and baselines, input and output sequence length: 720 (120 h). The best and second best performances in each dataset are in {\color{red}{red}} and {\color{blue}{blue}} colors.}
  \vspace{-0.1in}
  \label{tab:pred_perf}
  \begin{tabular}{@{} llcccccccccc @{}}
    \toprule
    Model & Metric & \multicolumn{2}{c}{agmar2020} & \multicolumn{2}{c}{agmar2021} & \multicolumn{2}{c}{agmar2022} & \multicolumn{2}{c}{agmar2023} & \multicolumn{2}{c}{agmar2024} \\
     &  & w/ WF & w/o WF & w/ WF & w/o WF & w/ WF & w/o WF & w/ WF & w/o WF & w/ WF & w/o WF \\

    \midrule
    MARLP (ours) & MSE ($\downarrow$) & \textbf{\color{red}0.331} & \textbf{\color{blue}0.579} & \textbf{\color{red}0.295} & \textbf{\color{red}0.351} & \textbf{\color{red}0.706} & 0.930 & \textbf{\color{red}0.309} & 0.421 & \textbf{\color{red}1.130} & 1.735 \\
    & MAE ($\downarrow$) & \textbf{\color{red}0.347} & \textbf{\color{blue}0.604} & \textbf{\color{red}0.428} & \textbf{\color{red}0.469} & \textbf{\color{red}0.657} & 0.792 & \textbf{\color{red}0.271} & \textbf{\color{red}0.376} & \textbf{\color{red}0.848} & 1.059 \\
    & peak\_time ($\downarrow$) & \textbf{\color{red}13.409} & \textbf{\color{red}29.750} & \textbf{\color{red}28.114} & \textbf{\color{red}37.063} & \textbf{\color{red}22.390} & 32.277 & \textbf{\color{red}10.967} & \textbf{\color{red}22.381} & \textbf{\color{red}44.092} & \textbf{\color{blue}50.385} \\
    & peak\_value ($\downarrow$) & \textbf{\color{red}0.723} & \textbf{\color{blue}1.116} & \textbf{\color{red}0.515} & 0.962 & \textbf{\color{red}1.034} & \textbf{\color{blue}1.610} & \textbf{\color{red}0.317} & \textbf{\color{red}0.521} & \textbf{\color{red}1.526} & 1.692 \\
    \midrule
    TimesNet\cite{wu2023timesnet} & MSE ($\downarrow$) & 0.576 & 0.603 & 0.527 & 0.483 & \textbf{\color{blue}0.883} & \textbf{\color{red}0.854} & 1.130 & 1.025 & 1.295 & \textbf{\color{red}1.149} \\
    & MAE ($\downarrow$) & 0.603 & 0.621 & 0.581 & 0.563 & \textbf{\color{blue}0.746} & \textbf{\color{red}0.725} & 0.924 & 0.868 & 0.935 & \textbf{\color{red}0.841} \\
    & peak\_time ($\downarrow$) & \textbf{\color{blue}33.060} & \textbf{\color{blue}32.522} & \textbf{\color{blue}35.670} & 45.182 & 30.537 & 42.625 & 40.228 & 44.371 & 70.404 & \textbf{\color{red}42.336} \\
    & peak\_value ($\downarrow$) & 1.448 & 1.327 & \textbf{\color{blue}0.789} & \textbf{\color{blue}0.938} & 1.789 & 1.763 & 0.984 & 1.126 & \textbf{\color{blue}1.528} & \textbf{\color{red}1.271} \\
    \midrule
    PatchTST/64\cite{nie2022time} & MSE ($\downarrow$) & \textbf{\color{blue}{0.465}} & \textbf{\color{red}0.490} & \textbf{\color{blue}0.337} & \textbf{\color{blue}0.374} & 1.545 & 0.864 & 0.342 & \textbf{\color{blue}0.399} & 1.311 & \textbf{\color{blue}1.480} \\
    & MAE ($\downarrow$) & \textbf{\color{blue}0.536} & \textbf{\color{red}0.547} & \textbf{\color{blue}0.456} & \textbf{\color{blue}0.483} & 0.988 & \textbf{\color{blue}0.736} & \textbf{\color{blue}0.431} & 0.466 & 0.913 & \textbf{\color{blue}0.989} \\
    & peak\_time ($\downarrow$) & 39.127 & 44.855 & 45.299 & 43.963 & \textbf{\color{blue}29.936} & \textbf{\color{red}26.776} & \textbf{\color{blue}28.925} & \textbf{\color{blue}25.388} & 55.352 & 61.752 \\
    & peak\_value ($\downarrow$) & \textbf{\color{blue}1.251} & 1.233 & 0.790 & \textbf{\color{red}0.704} & \textbf{\color{blue}1.728} & \textbf{\color{red}1.589} & \textbf{\color{blue}0.523} & \textbf{\color{blue}0.581} & 1.611 & \textbf{\color{blue}1.502} \\
    \midrule
    DLinear\cite{zeng2023transformers} & MSE ($\downarrow$) & 2.370 & 2.401 & 1.814 & 1.670 & 4.079 & 4.944 & \textbf{\color{blue}0.322} & \textbf{\color{red}0.327} & 1.598 & 1.730 \\
    & MAE ($\downarrow$) & 1.391 & 1.409 & 1.180 & 1.105 & 1.828 & 2.042 & 0.446 & \textbf{\color{blue}0.450} & 1.074 & 1.115 \\
    & peak\_time ($\downarrow$) & 74.776 & 74.501 & 42.604 & \textbf{\color{blue}38.959} & 56.322 & 55.657 & 43.076 & 41.139 & 50.871 & 52.143 \\
    & peak\_value ($\downarrow$) & 2.237 & 2.359 & 1.912 & 1.781 & 2.716 & 2.802 & 0.756 & 0.756 & 1.833 & 1.903 \\
    \midrule
    iTransformer\cite{liu2023itransformer} & MSE ($\downarrow$) & 0.537 & 0.621 & 0.387 & 0.565 & 0.864 & \textbf{\color{blue}0.855} & 0.477 & 1.749 & \textbf{\color{blue}1.151} & 1.732 \\
    & MAE ($\downarrow$) & 0.581 & 0.622 & 0.490 & 0.602 & 0.753 & 0.754 & 0.527 & 0.989 & \textbf{\color{blue}0.853} & 1.056 \\
    & peak\_time ($\downarrow$) & 53.088 & 41.893 & 47.908 & 44.291 & 27.952 & \textbf{\color{blue}30.450} & 29.978 & 39.140 & \textbf{\color{blue}46.569} & 51.133 \\
    & peak\_value ($\downarrow$) & 1.472 & \textbf{\color{red}0.938} & 1.066 & 0.773 & 1.885 & 1.639 & 0.776 & 1.214 & 1.535 & 1.703 \\
    \bottomrule
  \end{tabular}
\end{table*}
\section{In-Field Experiment Setup}\label{sec_deploy}
\begin{figure}[tp]
	{\includegraphics[width=.46\textwidth]{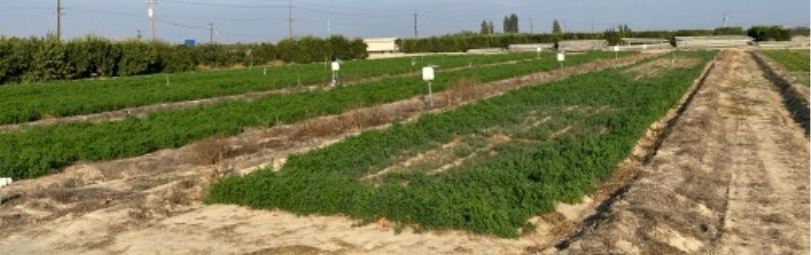}}
   	\vspace{-0.1in}
	\caption{The alfalfa field of 2023 experiments.}
	\label{fig_field_photo}
	\vspace{-0.1in}
\end{figure}

As shown in Figures~\ref{fig_field_photo} and \ref{fig_deploy_illus}, we build a real experimental testbed to verify the effectiveness of \ourSystem in practical scenarios.
It includes sensory data collection, transmission and decision making modules.
These components function as a comprehensive in-field experiment system that enables online evaluation of \ourSystem, offering more precise and realistic insights than those obtained from simulations or offline evaluations.

\textbf{Sensor Data Collection:} 
The oxygen and moisture sensors are placed at critical points throughout the facility. 
These sensors continuously measure real-time oxygen and moisture data, enabling prompt adjustments to maintain optimal oxygen levels.

\textbf{Sensor Data Transmission:} 
The collected oxygen and moisture data are transmitted to a central server through LoRa networks, ensuring long-range, low-power wireless communication~\cite{yang2024low, yang2023link, yang2024orchloc}. The implementation details are in Appendix~\ref{append:implement}.
We configured the transmission parameters to ensure that all measured sensory data are accurately received by our central server~\cite{yang2024rateless,  yang2022lldpc}.

\textbf{Inference Server.} 
This server is configured on the cloud~\cite{xu2024cloudeval} with an Intel(R) Core(TM) i9-11900KF CPU and an NVIDIA GEFORCE
RTX 3080 Ti GPU. It executes the MPC algorithm and forecasting neural networks with continuously aggregated sensor data and queried weather forecast data from open-source web API~\cite{openmeteo} in real time.

\textbf{System Overhead.} 
The power consumption of each sensor node for sensing and communicating is 64 mW, which can be easily covered by a solar panel, mitigating the need to change batteries. The gateway and server could be purchased as a service during flooding seasons. Overall, the system requires less than \$400 to implement and \$50/year to maintain, ensuring easy adoption, even for small farms.

\begin{figure}[tp]
	{\includegraphics[width=.48\textwidth]{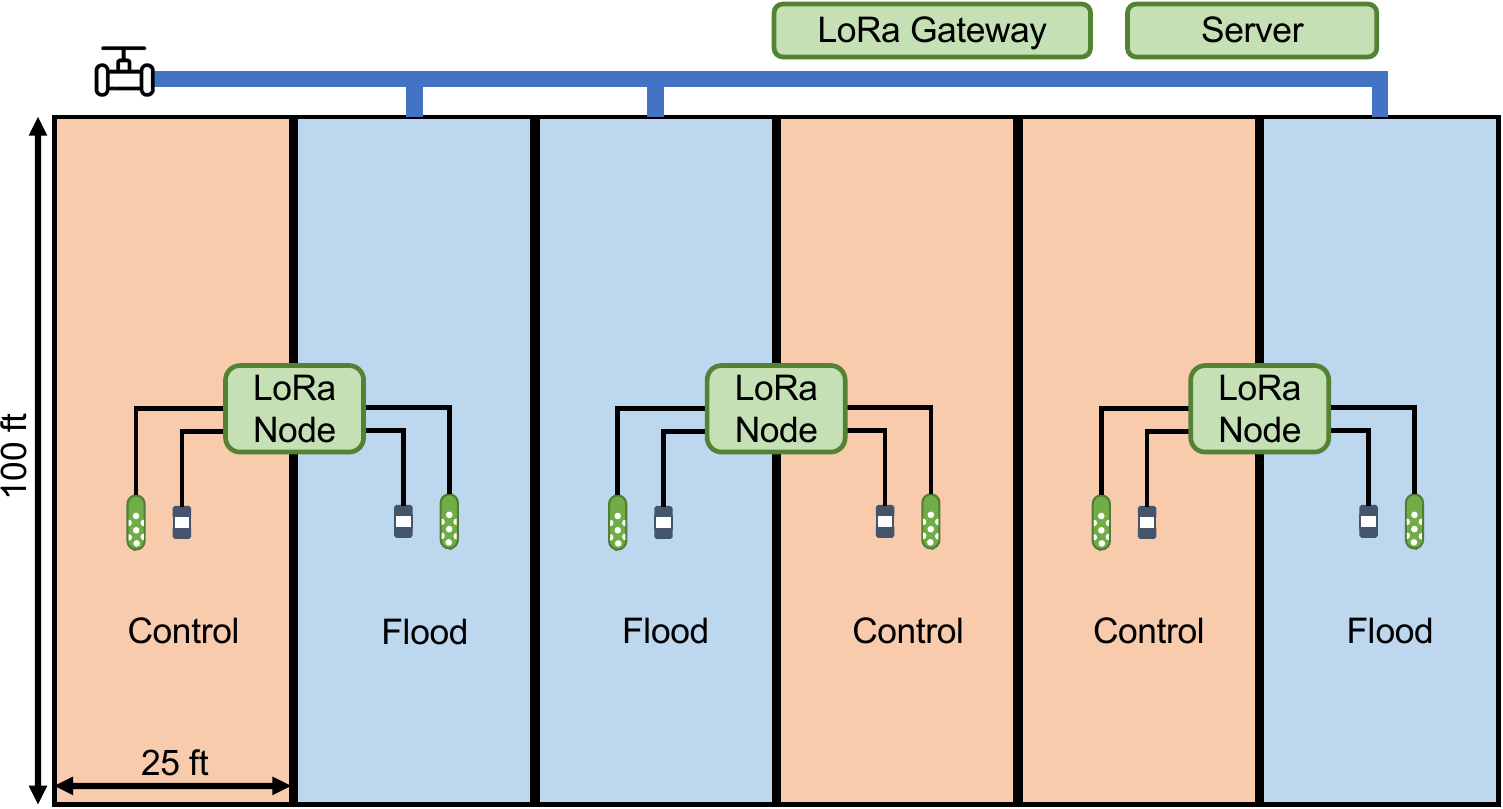}}
   	\vspace{-0.1in}
	\caption{The illustration of in-field deployment.}
	\label{fig_deploy_illus}
	\vspace{-0.1in}
\end{figure}

\section{Evaluation}

We ask the following questions to evaluate \ourSystem through in-field experiments and large-scale simulations:
\begin{enumerate}
\item[RQ1] How effective is \ourSystem in predicting oxygen curves?
\item[RQ2] How effective is \ourSystem in control performance?
\item[RQ3] Can \ourSystem be effectively generalized across different soil types, plant species, and weather patterns?
\item[RQ4] How effective is each design component of \ourSystem?
\end{enumerate}

\noindent To answer these questions, we evaluate the predictive capacity on datasets of past years in Section~\ref{sec_evaset} and assess the control performance of \ourSystem during the in-field deployment in Section~\ref{sec_infield}.
Then, we investigate the performance under different factors in Section~\ref{sec_simu}, followed by the ablation study in Section~\ref{sec_ablation}.

\subsection{Predictive Capability}\label{sec_evaset}

We choose four recent, representative and highly performed time-series forecasting models as baselines: TimesNet~\cite{wu2023timesnet}, PatchTST~\cite{nie2022time}, DLinear~\cite{zeng2023transformers}, and iTransformer~\cite{liu2023itransformer}, covering convolutional, linear, and transformer-based methodologies. 
We evaluate all baseline forecasting models on five datasets from five years within three fields.
The prediction sequence length is set as 720, which represents five days.
Given the sensor reading interval of 10 minutes, the forecasting window is 120 hours, which can fully observe oxygen recovery in most cases.
Table~\ref{tab:agmar_dataset} shows dataset statistics, including the collection area, period, flooding strategy, and sequence length. 
Ag-MAR actions lie between growing seasons, which is roughly from January to April for alfalfa in California, US.
From 2020 to 2023, the field is flooded with constant intervals, e.g., once a week. 
Trials in 2024 are controlled by \ourSystem.

Table~\ref{tab:pred_perf} reports the MSE, MAE, as well as the mean absolute error of the peak time (PTE) and peak value (PVE) among all forecasting models.
The unit of peak time error is an hour.
In Table~\ref{tab:pred_perf}, it is evident that \ourSystem consistently achieves the highest performance (highlighted with red values) when incorporating weather forecast data. 
This underscores the effectiveness of our long-term soil oxygen prediction algorithm,  which is vital for the MPC.

\begin{table}
\centering
\caption{The collection settings of Ag-MAR datasets.}
\vspace{-0.1in}
\label{tab:agmar_dataset}
\begin{tabular}{cccccc}
\toprule
Year & 2020 & 2021 & 2022 & 2023 & 2024 \\ \hline
Area($ft^2$) & 590*280 & 590*280 & 590*280 & 284*132 & 150*100\\
Flooding & Const. & Const. & Const. & Const. &  MARLP \\
Duration & 2/20-4/2 & 2/12-3/31 & 1/19-4/8 & 2/28-4/6 & 1/19-4/4 \\
Sequence & 6086 & 6902 & 11455 & 5389 & 11001 \\
\bottomrule
\end{tabular}
\end{table}

\subsection{In-field Control Experiments}\label{sec_infield}

We perform the first in-field deployment of the Ag-MAR control scheme, as illustrated in Section~\ref{sec_deploy}.
The quality of control is evaluated under two factors: \textbf{oxygen deficit ratio (ODR)} and \textbf{recharging amount}. 
ODR calculates the ratio of time that the soil oxygen level is below the safety threshold, which should be zero in ideal cases. 
The best recharging amount may vary according to weather conditions, so the optimal recharging amount can not be asserted, instead, we can only compare the schemes with each other. All in-field experiments are conducted in alfalfa fields, with the safety threshold of oxygen concentration as 10\%.

Table~\ref{tab:agmar_performance_comparison} compares the control performance of \ourSystem with the weekly flooded scheme in 2020-2023 as the baseline. 
The weekly flooded scheme resulted in an average oxygen deficit ratio of $2.72\%$, while controlling with \ourSystem yields an ODR of $0.36\%$. 
At the same time, \ourSystem increased the recharging amount (inch per week) from $7.64$ to $10.371$, with a $35.8\%$ improvement.

\begin{figure}[tp]
	\subfigure[Weekly flooding scheme fails to adapt to the weather conditions.]{
		\includegraphics[width=.225\textwidth]{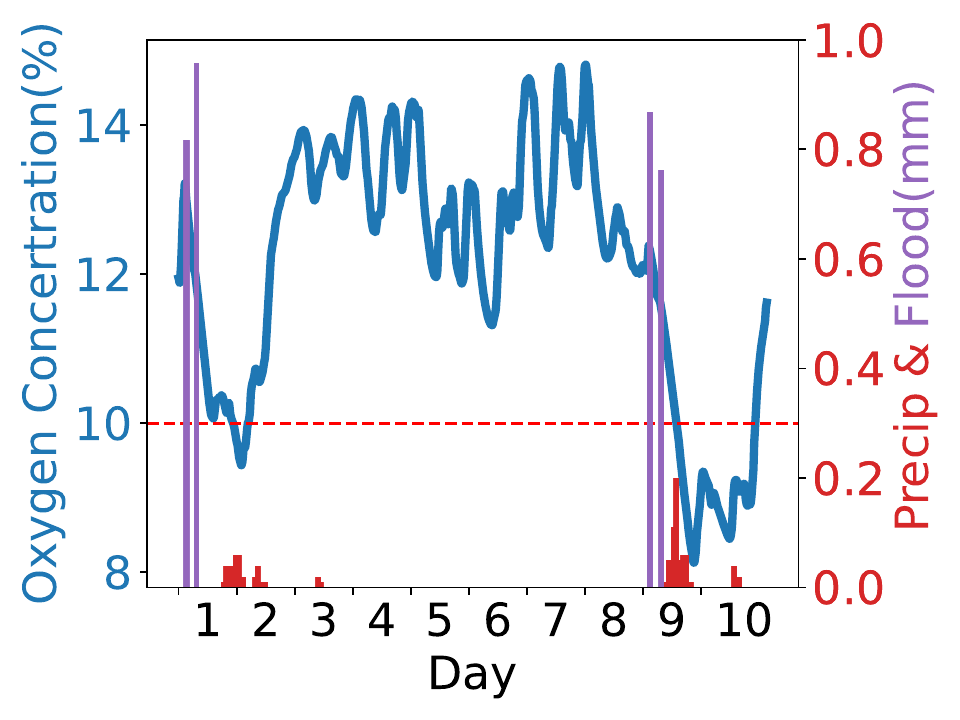}}
  \hspace{0.001in}
	\subfigure[\ourSystem can avoid potential oxygen deficits while grasping flooding chances.]{
		\includegraphics[width=.225\textwidth]{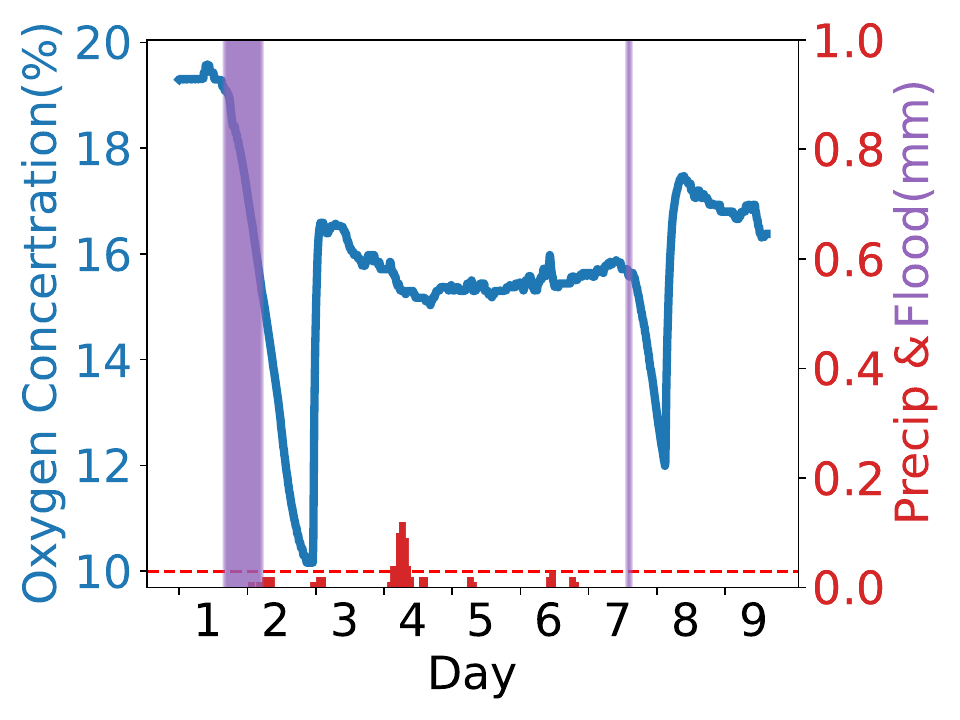}}
	\vspace{-0.1in}
 \caption{An example of how \ourSystem handles precipitation.}
	\label{fig_rain_case}
\end{figure}

\begin{table}
\centering
\caption{Control performance comparison.}
\vspace{-0.1in}
\label{tab:agmar_performance_comparison}
\begin{tabular}{ccc}
\toprule
& Const. & \ourSystem \\ \hline
Oxygen Deficit Ratio & 2.72\% & 0.36\%\\
Recharging Amount (inch per week) & 7.640 & 10.371 \\
\bottomrule
\end{tabular}
\vspace{-0.1in}
\end{table}

Figure~\ref{fig_rain_case} provides an example of comparison between the weekly flooding scheme and \ourSystem to show the reason behind the high effectiveness of \ourSystem. The red and purple bars represent the amount of precipitation and flooding, respectively. Each bar is 10 minutes wide, so the visual areas indicate the total amount of water input.
In Figure~\ref{fig_rain_case}~(a), the weekly-based approach ignores the heavy rain forecast after flooding, resulting in the unexpected oxygen deficit below 10\% on day 1 and day 8. 
At the same time, it misses the opportunity to flood during days 3-6, when there is no rainfall. 
Figure~\ref{fig_rain_case}~(b) shows the performance of \ourSystem. 
On day 1 and day 2, when the rainfall is negligible, it conducts a 14-hour recharge for 3.1 mm. The oxygen low peak is 10.16\%, without exceeding the safety threshold, as depicted by the red dashed line. 
Knowing the heavy rain on day 4 of the forecast, it discarded aggressive proposals and stopped flooding for 5 days to avoid danger. 
This shows that \ourSystem can foresee the consequences of each proposal and avoid risky flooding while making full use of flooding opportunities.

\subsection{Large-scale Simulation}\label{sec_simu}

To evaluate the generalization capability of all prediction models, we utilize a simulator to replicate a diverse array of factors, including soil types, crop species, and regional climate.\footnote{We didn't conduct in-field A/B tests due to the limits of field resources.}
Specifically, we simulate the water content transition in the soil using HYDRUS~\cite{vsimuunek1999hydrus}, a classical simulator for soil flux~\cite{bali2023use}.
Then we model the oxygen diffusion process, root respiration, and microbial respiration based on empirical equations~\cite{cook2003oxygen}. The evaluation targets the oxygen deficit ratio and recharging amount in the simulation.

\subsubsection{Impact of Soil Types}

\begin{figure}[tp]
	\subfigure[The oxygen deficit ratio.]{
		\includegraphics[width=.23\textwidth]{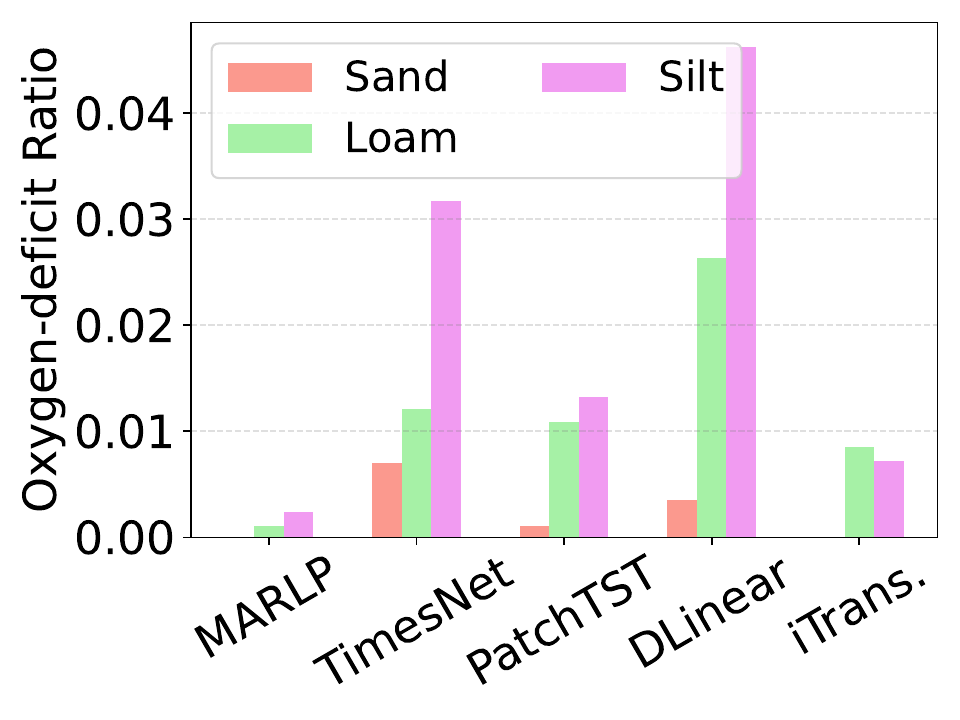}}
	\subfigure[The recharging amount.]{
		\includegraphics[width=.23\textwidth]{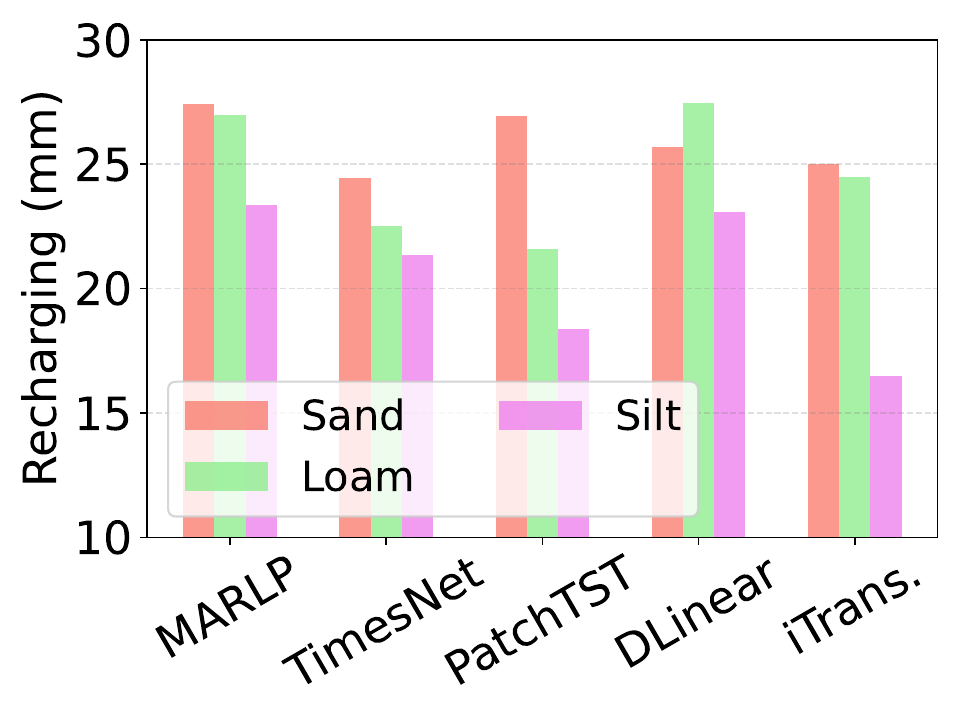}}
	\vspace{-0.1in}
 \caption{Control performance for soil types.}
	\label{fig_diff_soil}
	\vspace{-0.1in}
\end{figure}

To check the generalizability of \ourSystem on different soil, we mimic three representative soil types: sand, loam, and silt according to the soil texture triangle defined by USDA~\cite{usda1999united}.
We simulate each soil texture as a testbed, where all models are evaluated with MPC architecture the same as \ourSystem. The crops and weather remain the same as in the 2023 in-field experiment. 
Figure~\ref{fig_diff_soil}(a) shows that \ourSystem achieves the lowest oxygen deficit ratio than the baselines for all three soil textures, and Figure~\ref{fig_diff_soil}(b) shows that \ourSystem achieves a high recharging amount at the same time. 
Although DLinear achieves a higher recharging amount on loam and slit, it performs significantly worse than \ourSystem in terms of oxygen deficit ratio. The superior robustness based on the causality-aware forecasting of \ourSystem holds on other soil types.
Note that the soil texture has a significant impact on the general trend of the potential recharging amount because soil with relatively high percolation rates can accelerate the atmosphere exchange process.

\begin{figure}[tp]
	\subfigure[The oxygen deficit ratio.]{
		\includegraphics[width=.23\textwidth]{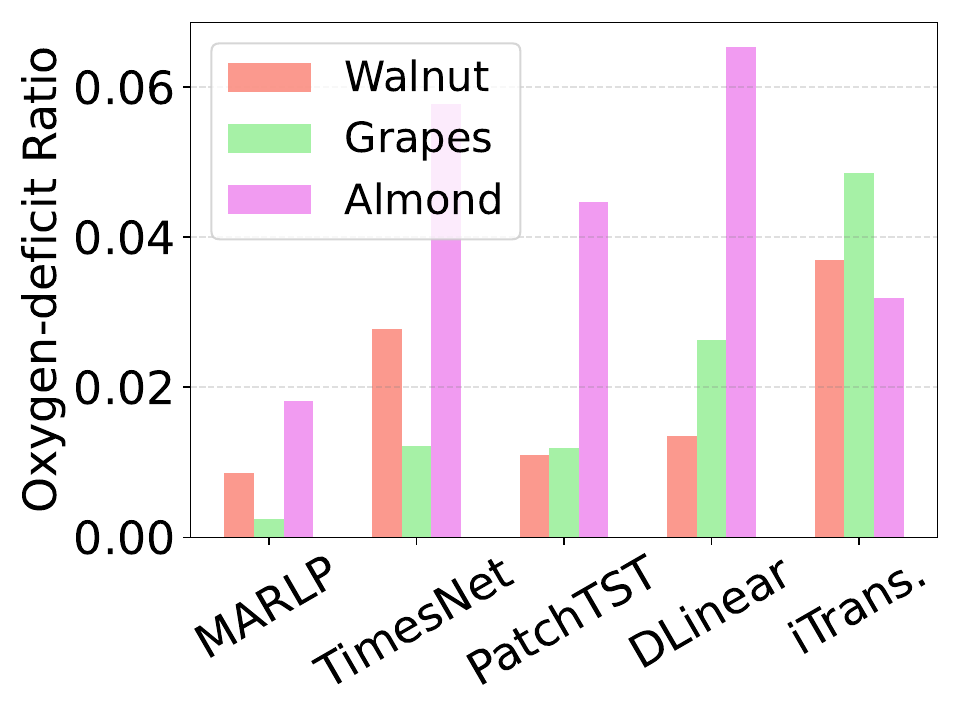}}
	\subfigure[The recharging amount.]{
		\includegraphics[width=.23\textwidth]{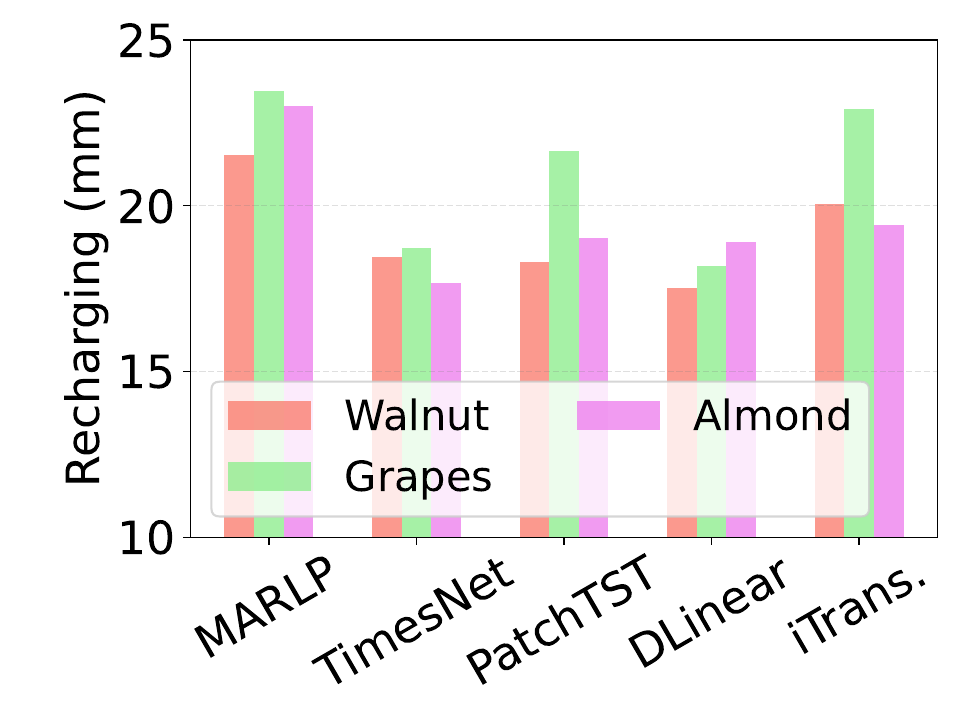}}
	\vspace{-0.1in}
 \caption{Control performance for crop species.}
	\label{fig_diff_plant}
	\vspace{-0.1in}
\end{figure}

\subsubsection{Impact of Crop Species}
We choose walnut trees, grapes and almond trees, with distinct root densities, depths and oxygen-deficit tolerances~\cite{o2015soil}, to assess how the crop diversity influences the control performance. 
Walnuts and almonds are less tolerant to flooding than alfalfa, hence they are more susceptible to high oxygen-deficit ratio. 
Figure~\ref{fig_diff_plant}(a) shows that \ourSystem achieves the lowest oxygen-deficit ratio than the baselines for all three crop species, and Figure~\ref{fig_diff_plant}(b) shows that \ourSystem achieves a high recharging amount at the same time. Therefore, \ourSystem has the best generalization capability among all methods.

\subsubsection{Impact of Regional Climatic}
\begin{figure}[tp]
	\subfigure[The oxygen deficit ratio.]{
		\includegraphics[width=.23\textwidth]{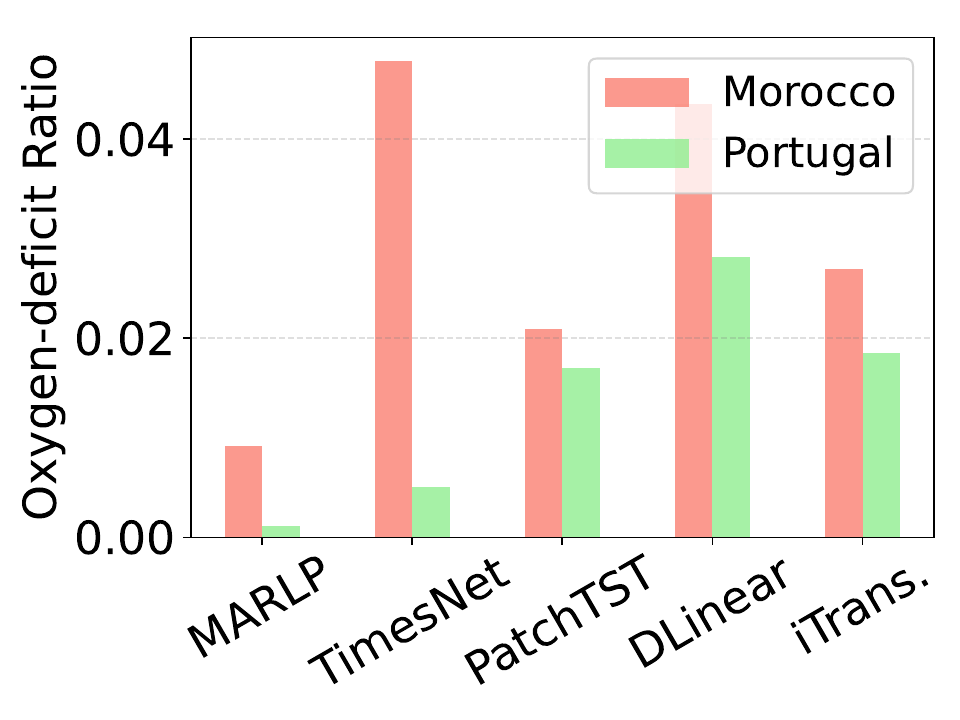}}
	\subfigure[The recharging amount.]{
		\includegraphics[width=.23\textwidth]{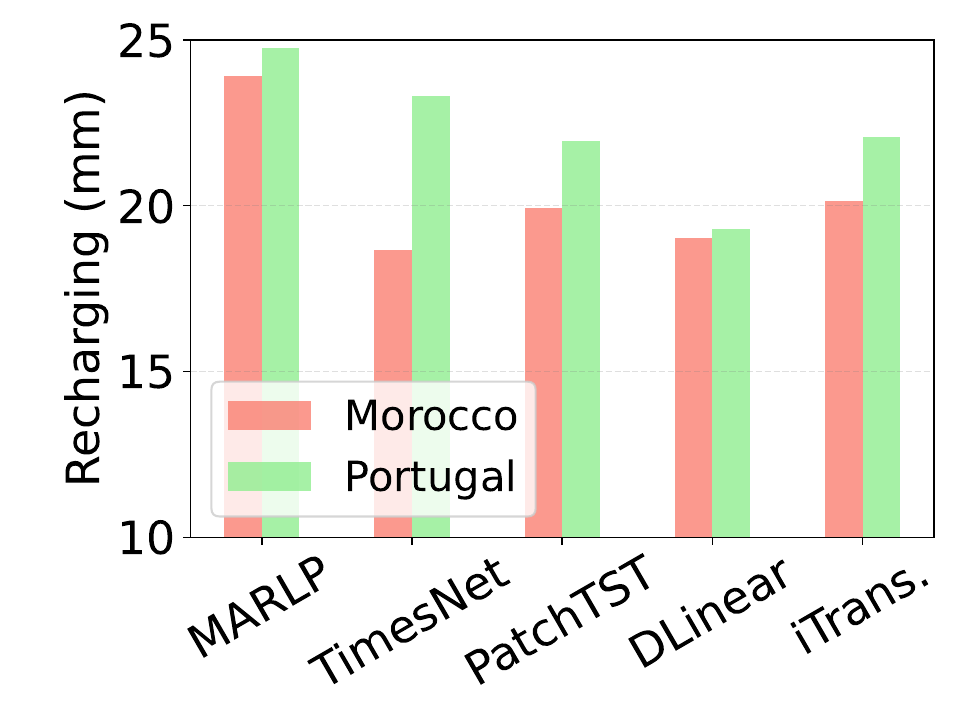}}
	\vspace{-0.1in}
 \caption{Control performance for climatic features in different regions.}
	\label{fig_diff_region}
	\vspace{-0.1in}
\end{figure}

Although California is at the forefront of Ag-MAR adoption, this practice is becoming increasingly popular in other parts of the world that face similar hydrological challenges, particularly those with a Mediterranean climate. 
We broaden its scope to include the High-Atlas region in Morocco~\cite{bouimouass2024importance} and the Algarve region in Portugal~\cite{standen2023integration}, incorporating soil and weather data from these diverse global regions that are actively exploring Ag-MAR.
The experimental periods for the simulations are both Feb. 28th - Apr. 6th, 2023, aligning with our real-world experiments in California. We use HYDRUS~\cite{vsimuunek1999hydrus} to simulate the soil dynamics of these two sites and use five models to conduct flooding control separately, with the same goal settings.
The results are shown in Figure~\ref{fig_diff_region}. The control performance of all models exhibits drops since the High-Atlas encounters a few sharp precipitations due to high altitude, while Algarve is located by the sea, with stronger daily periodicity. 
\ourSystem keeps achieving the best control performance. 
It underscores the generalizability of \ourSystem and its potential to be adapted worldwide.

\subsection{Ablation Study}\label{sec_ablation}

In this section, we systematically tested the performance of \ourSystem by excluding each design module to evaluate their individual effectiveness. 
Additionally, we examined how each input variable contributes to the system. 
These tests revealed the necessity of involving these clues and casual strength between the soil oxygen and each of them. 
We plot the mean and standard deviation of MSE across all datasets.

\begin{figure}[tp]
	\subfigure[Ablation study for modules.]{
		\includegraphics[width=.23\textwidth]{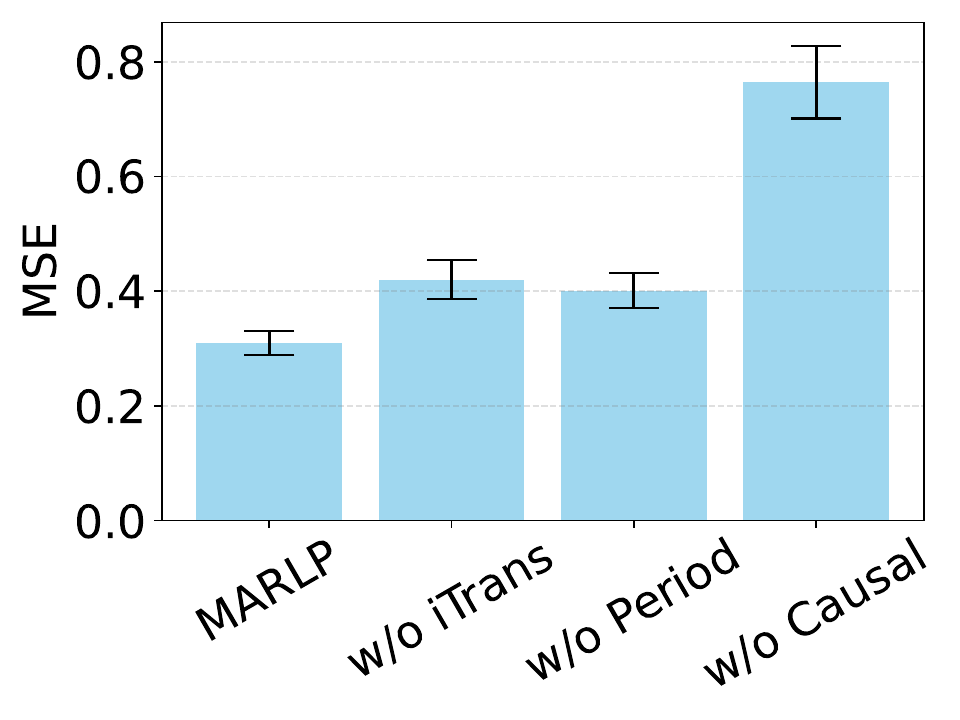}}
	\subfigure[Ablation study for input variables.]{
		\includegraphics[width=.23\textwidth]{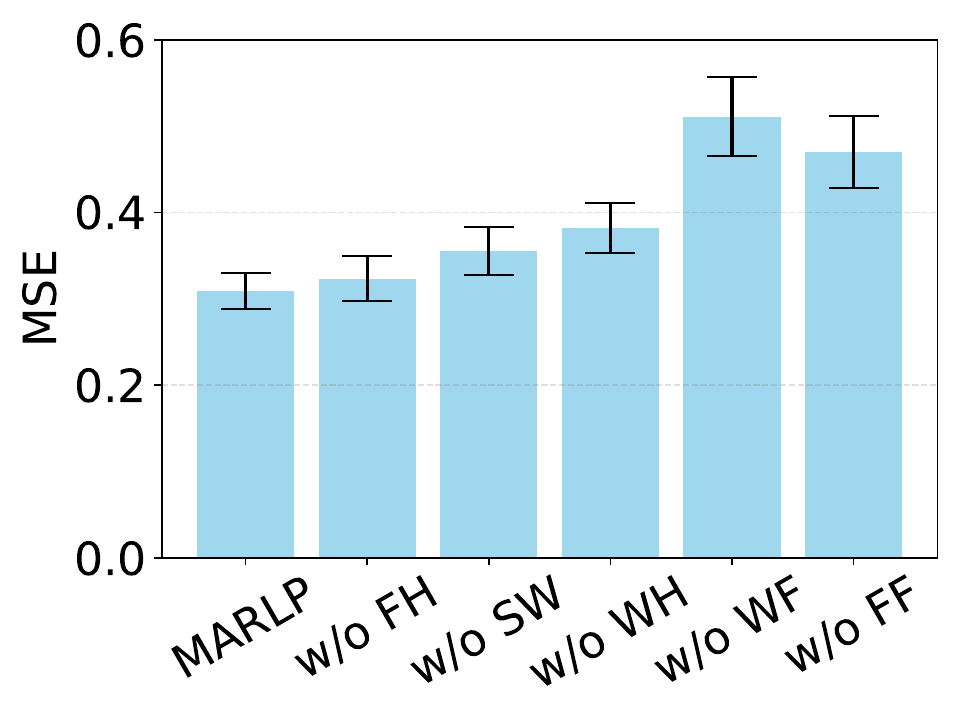}}
	\vspace{-0.1in}
 \caption{Ablation study for modules and input variables.}
	\label{fig_ablation}
	\vspace{-0.1in}
\end{figure}

\textbf{Design module ablations.}
We construct ablated versions of the model by substituting the iTransformer with a vanilla Transformer, removing the periodicity block and the causal module.
Figure~\ref{fig_ablation}(a) illustrates that the performance of each ablated version diminishes, with the version lacking the causal module experiencing the most significant reduction.
This decline is because of the pronounced causality inherent in Ag-MAR data.

\textbf{Impact of Input Features.}
Figure~\ref{fig_ablation}(b) shows the influence of various inputs on the MSE of oxygen forecasting.
The abbreviations FH, SW, WH, WF, and FF represent flooding history, soil water content, weather history, weather forecast, and future flooding, respectively. 
All inputs play a pivotal role in enhancing the final performance, with weather forecasts and future flooding making the most significant contributions. 
Consequently, optimal forecasting necessitates a synergistic integration of historical data.

Overall, this ablation study affirms the importance of each factor in improving oxygen forecasting and underscores the significance of causal-aware forecasting that incorporates external indicators.

\vspace{-0.05in}
\section{Discussions and Future Works}

\textbf{Ag-MAR studies. }  
Ag-MAR has been studied in terms of environmental benefits and potential risks\cite{bali2023use, ganot2021natural, kourakos2019increasing}, the fitness of different soil types, crops, and geological regions. The preliminary results are positive and bring it towards global adoption~\cite{marwaha2021identifying}. However, its current implementation is empirical~\cite{ganot2021model}, lacking a standardized and automatic workflow to achieve the best flooding. Based on the oxygen pattern analysis, \ourSystem provides a systematical solution that can work seamlessly with sensor systems, without the need for expert knowledge. 
Recent studies also illustrate that the oxygen-deficit tolerance level is temperature dependent~\cite{barta2002interaction}. Future works can integrate this feature to enhance the practicality of \ourSystem.

\textbf{Multi-variate time-series prediction algorithms. } 
The evolution of multivariate time series prediction algorithms has been marked by significant milestones, starting with the development of the autoregressive integrated moving average (ARIMA) model~\cite{box1968some}, progressing through the use of recurring neural networks (RNN)~\cite{hochreiter1997long,cho2014learning}, and further evolving with the introduction of Transformer~\cite{vaswani2017attention}. 
NS-Transformer~\cite{liu2022non} proposes series stationarization and de-stationary attention to handle the distribution shift and boost the performance of the Transformer on time-series prediction. 
Dish-TS~\cite{fan2023dish} proposes a general paradigm to alleviate distribution shifts in time series.
ESG~\cite{ye2022learning} integrates learning pairwise correlations and temporal dependency in one framework.
iTransformer~\cite{liu2023itransformer} optimizes the extraction of temporal and cross-variate dependency by swapping two modules. Instead of conducting universal forecasting, other works focus on different objectives. 
TimesNet~\cite{wu2023timesnet} and DEPTS~\cite{fan2021depts} aims to learn periodic patterns in time series.
TSMixer~\cite{10.1145/3580305.3599533} utilizes the MLP architecture to reduce computing overhead. 
IPOC~\cite{chen2023ipoc} innovates with ensemble learning, offering real-time adaptable confidence interval predictions.
\ourSystem distinguishes itself with a two-stage method that first predicts based on historical patterns, then utilizes external clues for causality-aware calibration.

\textbf{Causal discovery and inference for time-series data.} Causal relationship widely exists in sequence data, e.g., the classical causality between milk price and butter price~\cite{awokuse2009threshold}. Granger causality is proposed to handle the delay of the causal impact~\cite{seth2007granger}. \cite{huang2019causal} proposes Bayesian forecasting with time-varying causal models, but it only works for short terms. CASTOR~\cite{rahmani2023castor} introduces time-lagged links into GNN to enhance Granger causality modeling. REASON~\cite{wang2023interdependent} utilizes graph neural networks to extract layered causal relationships. CORAL~\cite{wang2023incremental} is a framework that automatically updates the root cause analysis model. Instead of discovering and quantifying causality, our solution focuses on applying causality to long-term time-series forecasting with external clues.

\textbf{Time-series forecasting for real applications. } 
Time-series forecasting is not only a critical question in agricultural practices, but also vital in other industries such as electricity, weather, finance, traffic, and human-computer interaction~\cite{ren2021rapt, verma2023climode, liu2022leveraging, liu2021video, liu2023financial}. For example, RAPT~\cite{ren2021rapt} performs a prediction on medical data for the diagnosis of pregnancy complications. ClimODE~\cite{verma2023climode} uses physics-informed neural ODE to simulate global atmosphere dynamics for climate forecasting. We will upgrade the sensor systems to enable neural ODE on Ag-MAR oxygen modeling in future works.

\textbf{MPC based on long-term forecasting. } MPC has been used for application scenarios from the microsecond-level horizon (e.g., embedded system voltage control~\cite{liegmann2021real}) to the hour-level horizon (e.g., building control~\cite{an2023clue, an2024go}, grid control~\cite{nelson2020model}). Existing works utilize MPC for irrigation, but the planning horizon and forecast window are less than one day~\cite{ding2022drlic, ding2023optimizing}. 
Reinforcement learning \cite{ding2023exploring, shen2019deepapp, ding2023multi} based control can achieve effective control, but lacks reliability and requires a large amount of data to converge, making it unsuitable for Ag-MAR. 
To the best of our knowledge, this work is the first MPC work to consider a planning horizon of several days.

\textbf{Robustness and Efficiency.} 
Robustness is a major concern for applied machine learning~\cite{phatak2023computing, raghvendra2024new, lahn2024combinatorial}. The predicted state trajectory generated as part of the MPC planning process allows the safety check of the trajectory, which offers a higher level of reliability~\cite{an2024reward}. 
In future works, we may apply safe guarantee mechanisms like a Gaussian-based uncertainty check[5], to achieve better tradeoffs between reliability and efficiency.
Furthermore, given the typically extensive search spaces involved in real-world, long-term forecasting control, efficiency becomes a pivotal aspect. Unlike RL, which can separate agent models from environmental models to improve efficiency~\cite{pmlr-v202-gmelin23a, lan2024improved, lan2024asynchronous}, time series modeling offers limited scope to balance performance with efficiency. However, recent advances in state space models (SSM) have significantly mitigated computational burdens~\cite{gu2023mamba}. Future research could explore the application of SSMs in scenarios where real-time requirements are critical.


\section{Conclusion}

This paper introduces \ourSystem, a model predictive control system for Ag-MAR, employing heuristic planning and a long-term oxygen forecasting module to optimize groundwater recharge and soil oxygen levels. Benefiting from a causality-aware forecasting model, \ourSystem effectively manages environmental variables in real-time, enhancing water use efficiency in agriculture. The successful deployment of \ourSystem, which reduces the oxygen deficit ratio and improves the total amount of applied water, showcases the system's potential in precision agriculture and sustainable resource management. 

\section*{Acknowledgments}
This work was supported in part by a UC Merced Fall 2023 Climate Action Seed Competition grant, and a UC Merced Spring 2023 Climate Action Seed Competition grant. 
Kang Yang was supported by a financial assistance award approved by the Economic Development Administration's Farms Food Future program.
Any opinions, findings, and conclusions expressed in this material are those of the authors and do not necessarily reflect the views of the funding agencies.

\balance

\bibliographystyle{ACM-Reference-Format.bst}
\bibliography{9_references}


\begin{thebibliography}{73}


\ifx \showCODEN    \undefined \def \showCODEN     #1{\unskip}     \fi
\ifx \showDOI      \undefined \def \showDOI       #1{#1}\fi
\ifx \showISBNx    \undefined \def \showISBNx     #1{\unskip}     \fi
\ifx \showISBNxiii \undefined \def \showISBNxiii  #1{\unskip}     \fi
\ifx \showISSN     \undefined \def \showISSN      #1{\unskip}     \fi
\ifx \showLCCN     \undefined \def \showLCCN      #1{\unskip}     \fi
\ifx \shownote     \undefined \def \shownote      #1{#1}          \fi
\ifx \showarticletitle \undefined \def \showarticletitle #1{#1}   \fi
\ifx \showURL      \undefined \def \showURL       {\relax}        \fi
\providecommand\bibfield[2]{#2}
\providecommand\bibinfo[2]{#2}
\providecommand\natexlab[1]{#1}
\providecommand\showeprint[2][]{arXiv:#2}

\bibitem[Amos et~al\mbox{.}(2018)]%
        {amos2018differentiable}
\bibfield{author}{\bibinfo{person}{Brandon Amos}, \bibinfo{person}{Ivan Jimenez}, \bibinfo{person}{Jacob Sacks}, \bibinfo{person}{Byron Boots}, {and} \bibinfo{person}{J~Zico Kolter}.} \bibinfo{year}{2018}\natexlab{}.
\newblock \showarticletitle{Differentiable mpc for end-to-end planning and control}.
\newblock \bibinfo{journal}{\emph{Advances in neural information processing systems}}  \bibinfo{volume}{31} (\bibinfo{year}{2018}).
\newblock


\bibitem[An et~al\mbox{.}(2024a)]%
        {an2024go}
\bibfield{author}{\bibinfo{person}{Zhiyu An}, \bibinfo{person}{Xianzhong Ding}, {and} \bibinfo{person}{Wan Du}.} \bibinfo{year}{2024}\natexlab{a}.
\newblock \showarticletitle{{Go Beyond Black-box Policies: Rethinking the Design of Learning Agent for Interpretable and Verifiable HVAC Control}}.
\newblock \bibinfo{journal}{\emph{arXiv preprint arXiv:2403.00172}} (\bibinfo{year}{2024}).
\newblock


\bibitem[An et~al\mbox{.}(2024b)]%
        {an2024reward}
\bibfield{author}{\bibinfo{person}{Zhiyu An}, \bibinfo{person}{Xianzhong Ding}, {and} \bibinfo{person}{Wan Du}.} \bibinfo{year}{2024}\natexlab{b}.
\newblock \showarticletitle{Reward Bound for Behavioral Guarantee of Model-based Planning Agents}.
\newblock \bibinfo{journal}{\emph{arXiv preprint arXiv:2402.13419}} (\bibinfo{year}{2024}).
\newblock


\bibitem[An et~al\mbox{.}(2023)]%
        {an2023clue}
\bibfield{author}{\bibinfo{person}{Zhiyu An}, \bibinfo{person}{Xianzhong Ding}, \bibinfo{person}{Arya Rathee}, {and} \bibinfo{person}{Wan Du}.} \bibinfo{year}{2023}\natexlab{}.
\newblock \showarticletitle{{CLUE: Safe Model-Based RL HVAC Control Using Epistemic Uncertainty Estimation}}. In \bibinfo{booktitle}{\emph{ACM BuildSys}}.
\newblock


\bibitem[Awokuse and Wang(2009)]%
        {awokuse2009threshold}
\bibfield{author}{\bibinfo{person}{Titus~O Awokuse} {and} \bibinfo{person}{Xiaohong Wang}.} \bibinfo{year}{2009}\natexlab{}.
\newblock \showarticletitle{Threshold effects and asymmetric price adjustments in US dairy markets}.
\newblock \bibinfo{journal}{\emph{Canadian Journal of Agricultural Economics/Revue canadienne d'agroeconomie}} \bibinfo{volume}{57}, \bibinfo{number}{2} (\bibinfo{year}{2009}), \bibinfo{pages}{269--286}.
\newblock


\bibitem[Bali et~al\mbox{.}(2023)]%
        {bali2023use}
\bibfield{author}{\bibinfo{person}{Khaled~M Bali}, \bibinfo{person}{Abdelmoneim~Zakaria Mohamed}, \bibinfo{person}{Sultan Begna}, \bibinfo{person}{Dong Wang}, \bibinfo{person}{Daniel Putnam}, \bibinfo{person}{Helen~E Dahlke}, {and} \bibinfo{person}{Mohamed~Galal Eltarabily}.} \bibinfo{year}{2023}\natexlab{}.
\newblock \showarticletitle{The use of HYDRUS-2D to simulate intermittent Agricultural Managed Aquifer Recharge (Ag-MAR) in Alfalfa in the San Joaquin Valley}.
\newblock \bibinfo{journal}{\emph{Agricultural Water Management}}  \bibinfo{volume}{282} (\bibinfo{year}{2023}), \bibinfo{pages}{108296}.
\newblock


\bibitem[Barta and Sulc(2002)]%
        {barta2002interaction}
\bibfield{author}{\bibinfo{person}{AL Barta} {and} \bibinfo{person}{RM Sulc}.} \bibinfo{year}{2002}\natexlab{}.
\newblock \showarticletitle{Interaction between waterlogging injury and irradiance level in alfalfa}.
\newblock \bibinfo{journal}{\emph{Crop science}} \bibinfo{volume}{42}, \bibinfo{number}{5} (\bibinfo{year}{2002}), \bibinfo{pages}{1529--1534}.
\newblock


\bibitem[Bouimouass et~al\mbox{.}(2024)]%
        {bouimouass2024importance}
\bibfield{author}{\bibinfo{person}{Houssne Bouimouass}, \bibinfo{person}{Sarah Tweed}, \bibinfo{person}{Vincent Marc}, \bibinfo{person}{Younes Fakir}, \bibinfo{person}{Hamza Sahraoui}, {and} \bibinfo{person}{Marc Leblanc}.} \bibinfo{year}{2024}\natexlab{}.
\newblock \showarticletitle{The importance of mountain-block recharge in semiarid basins: An insight from the High-Atlas, Morocco}.
\newblock \bibinfo{journal}{\emph{Journal of Hydrology}} (\bibinfo{year}{2024}), \bibinfo{pages}{130818}.
\newblock


\bibitem[Box and Jenkins(1968)]%
        {box1968some}
\bibfield{author}{\bibinfo{person}{George~EP Box} {and} \bibinfo{person}{Gwilym~M Jenkins}.} \bibinfo{year}{1968}\natexlab{}.
\newblock \showarticletitle{Some recent advances in forecasting and control}.
\newblock \bibinfo{journal}{\emph{Journal of the Royal Statistical Society. Series C (Applied Statistics)}} \bibinfo{volume}{17}, \bibinfo{number}{2} (\bibinfo{year}{1968}), \bibinfo{pages}{91--109}.
\newblock


\bibitem[{California Department of Water Resources}(2014)]%
        {Sgma}
\bibfield{author}{\bibinfo{person}{{California Department of Water Resources}}.} \bibinfo{year}{2014}\natexlab{}.
\newblock \bibinfo{title}{{Sustainable Groundwater Management Act (SGMA)}}.
\newblock \bibinfo{howpublished}{\url{https://water.ca.gov/programs/groundwater-management/sgma-groundwater-management}}.
\newblock


\bibitem[Chen et~al\mbox{.}(2023)]%
        {chen2023ipoc}
\bibfield{author}{\bibinfo{person}{Jiadong Chen}, \bibinfo{person}{Yang Luo}, \bibinfo{person}{Xiuqi Huang}, \bibinfo{person}{Fuxin Jiang}, \bibinfo{person}{Yangguang Shi}, \bibinfo{person}{Tieying Zhang}, {and} \bibinfo{person}{Xiaofeng Gao}.} \bibinfo{year}{2023}\natexlab{}.
\newblock \showarticletitle{IPOC: An Adaptive Interval Prediction Model based on Online Chasing and Conformal Inference for Large-Scale Systems}. In \bibinfo{booktitle}{\emph{Proceedings of the 29th ACM SIGKDD Conference on Knowledge Discovery and Data Mining}}. \bibinfo{pages}{202--212}.
\newblock


\bibitem[Cho et~al\mbox{.}(2014)]%
        {cho2014learning}
\bibfield{author}{\bibinfo{person}{Kyunghyun Cho}, \bibinfo{person}{Bart Van~Merri{\"e}nboer}, \bibinfo{person}{Caglar Gulcehre}, \bibinfo{person}{Dzmitry Bahdanau}, \bibinfo{person}{Fethi Bougares}, \bibinfo{person}{Holger Schwenk}, {and} \bibinfo{person}{Yoshua Bengio}.} \bibinfo{year}{2014}\natexlab{}.
\newblock \showarticletitle{Learning phrase representations using RNN encoder-decoder for statistical machine translation}.
\newblock \bibinfo{journal}{\emph{arXiv preprint arXiv:1406.1078}} (\bibinfo{year}{2014}).
\newblock


\bibitem[Cook and Knight(2003)]%
        {cook2003oxygen}
\bibfield{author}{\bibinfo{person}{FJ Cook} {and} \bibinfo{person}{JH Knight}.} \bibinfo{year}{2003}\natexlab{}.
\newblock \showarticletitle{Oxygen transport to plant roots: modeling for physical understanding of soil aeration}.
\newblock \bibinfo{journal}{\emph{Soil Science Society of America Journal}} \bibinfo{volume}{67}, \bibinfo{number}{1} (\bibinfo{year}{2003}), \bibinfo{pages}{20--31}.
\newblock


\bibitem[Dahlke et~al\mbox{.}(2018)]%
        {dahlke2018managed}
\bibfield{author}{\bibinfo{person}{Helen~E Dahlke}, \bibinfo{person}{Andrew~G Brown}, \bibinfo{person}{Steve Orloff}, \bibinfo{person}{Daniel~H Putnam}, {and} \bibinfo{person}{Toby O'Geen}.} \bibinfo{year}{2018}\natexlab{}.
\newblock \showarticletitle{Managed winter flooding of alfalfa recharges groundwater with minimal crop damage}.
\newblock \bibinfo{journal}{\emph{California Agriculture}} \bibinfo{volume}{72}, \bibinfo{number}{1} (\bibinfo{year}{2018}).
\newblock


\bibitem[Ding et~al\mbox{.}(2023a)]%
        {ding2023exploring}
\bibfield{author}{\bibinfo{person}{Xianzhong Ding}, \bibinfo{person}{Alberto Cerpa}, {and} \bibinfo{person}{Wan Du}.} \bibinfo{year}{2023}\natexlab{a}.
\newblock \showarticletitle{Exploring deep reinforcement learning for holistic smart building control}.
\newblock \bibinfo{journal}{\emph{ACM Transactions on Sensor Networks}} (\bibinfo{year}{2023}).
\newblock


\bibitem[Ding et~al\mbox{.}(2023b)]%
        {ding2023multi}
\bibfield{author}{\bibinfo{person}{Xianzhong Ding}, \bibinfo{person}{Alberto Cerpa}, {and} \bibinfo{person}{Wan Du}.} \bibinfo{year}{2023}\natexlab{b}.
\newblock \showarticletitle{Multi-zone HVAC Control with Model-Based Deep Reinforcement Learning}.
\newblock \bibinfo{journal}{\emph{arXiv preprint arXiv:2302.00725}} (\bibinfo{year}{2023}).
\newblock


\bibitem[Ding and Du(2022)]%
        {ding2022drlic}
\bibfield{author}{\bibinfo{person}{Xianzhong Ding} {and} \bibinfo{person}{Wan Du}.} \bibinfo{year}{2022}\natexlab{}.
\newblock \showarticletitle{{DRLIC: Deep Reinforcement Learning for Irrigation Control}}. In \bibinfo{booktitle}{\emph{ACM/IEEE IPSN}}.
\newblock


\bibitem[Ding and Du(2023)]%
        {ding2023optimizing}
\bibfield{author}{\bibinfo{person}{Xianzhong Ding} {and} \bibinfo{person}{Wan Du}.} \bibinfo{year}{2023}\natexlab{}.
\newblock \showarticletitle{Optimizing irrigation efficiency using deep reinforcement learning in the field}.
\newblock \bibinfo{journal}{\emph{ACM Transactions on Sensor Networks}} (\bibinfo{year}{2023}).
\newblock


\bibitem[Ekambaram et~al\mbox{.}(2023)]%
        {10.1145/3580305.3599533}
\bibfield{author}{\bibinfo{person}{Vijay Ekambaram}, \bibinfo{person}{Arindam Jati}, \bibinfo{person}{Nam Nguyen}, \bibinfo{person}{Phanwadee Sinthong}, {and} \bibinfo{person}{Jayant Kalagnanam}.} \bibinfo{year}{2023}\natexlab{}.
\newblock \showarticletitle{TSMixer: Lightweight MLP-Mixer Model for Multivariate Time Series Forecasting}. In \bibinfo{booktitle}{\emph{Proceedings of the 29th ACM SIGKDD Conference on Knowledge Discovery and Data Mining}} \emph{(\bibinfo{series}{KDD '23})}. \bibinfo{pages}{459–469}.
\newblock


\bibitem[Escriva-Bou et~al\mbox{.}(2017)]%
        {escriva2017water}
\bibfield{author}{\bibinfo{person}{Alvar Escriva-Bou}, \bibinfo{person}{Brian Gray}, \bibinfo{person}{Sarge Green}, \bibinfo{person}{Thomas Harter}, \bibinfo{person}{Richard Howitt}, \bibinfo{person}{Duncan MacEwan}, {and} \bibinfo{person}{N Seavy}.} \bibinfo{year}{2017}\natexlab{}.
\newblock \showarticletitle{Water Stress and a Changing San Joaquin Valley}.
\newblock \bibinfo{journal}{\emph{Public Policy Institute of California. https://www. ppic. org/content/pubs/report/R\_0317EHR. pdf}} (\bibinfo{year}{2017}).
\newblock


\bibitem[Fan et~al\mbox{.}(2023)]%
        {fan2023dish}
\bibfield{author}{\bibinfo{person}{Wei Fan}, \bibinfo{person}{Pengyang Wang}, \bibinfo{person}{Dongkun Wang}, \bibinfo{person}{Dongjie Wang}, \bibinfo{person}{Yuanchun Zhou}, {and} \bibinfo{person}{Yanjie Fu}.} \bibinfo{year}{2023}\natexlab{}.
\newblock \showarticletitle{Dish-ts: a general paradigm for alleviating distribution shift in time series forecasting}. In \bibinfo{booktitle}{\emph{Proceedings of the AAAI Conference on Artificial Intelligence}}.
\newblock


\bibitem[Fan et~al\mbox{.}(2021)]%
        {fan2021depts}
\bibfield{author}{\bibinfo{person}{Wei Fan}, \bibinfo{person}{Shun Zheng}, \bibinfo{person}{Xiaohan Yi}, \bibinfo{person}{Wei Cao}, \bibinfo{person}{Yanjie Fu}, \bibinfo{person}{Jiang Bian}, {and} \bibinfo{person}{Tie-Yan Liu}.} \bibinfo{year}{2021}\natexlab{}.
\newblock \showarticletitle{DEPTS: Deep Expansion Learning for Periodic Time Series Forecasting}. In \bibinfo{booktitle}{\emph{International Conference on Learning Representations}}.
\newblock


\bibitem[Ganot and Dahlke(2021a)]%
        {ganot2021model}
\bibfield{author}{\bibinfo{person}{Yonatan Ganot} {and} \bibinfo{person}{Helen~E Dahlke}.} \bibinfo{year}{2021}\natexlab{a}.
\newblock \showarticletitle{A model for estimating Ag-MAR flooding duration based on crop tolerance, root depth, and soil texture data}.
\newblock \bibinfo{journal}{\emph{Agricultural Water Management}}  \bibinfo{volume}{255} (\bibinfo{year}{2021}), \bibinfo{pages}{107031}.
\newblock


\bibitem[Ganot and Dahlke(2021b)]%
        {ganot2021natural}
\bibfield{author}{\bibinfo{person}{Yonatan Ganot} {and} \bibinfo{person}{Helen~E Dahlke}.} \bibinfo{year}{2021}\natexlab{b}.
\newblock \showarticletitle{Natural and forced soil aeration during agricultural managed aquifer recharge}.
\newblock \bibinfo{journal}{\emph{Vadose Zone Journal}} \bibinfo{volume}{20}, \bibinfo{number}{3} (\bibinfo{year}{2021}), \bibinfo{pages}{e20128}.
\newblock


\bibitem[Gmelin et~al\mbox{.}(2023)]%
        {pmlr-v202-gmelin23a}
\bibfield{author}{\bibinfo{person}{Kevin Gmelin}, \bibinfo{person}{Shikhar Bahl}, \bibinfo{person}{Russell Mendonca}, {and} \bibinfo{person}{Deepak Pathak}.} \bibinfo{year}{2023}\natexlab{}.
\newblock \showarticletitle{Efficient {RL} via Disentangled Environment and Agent Representations}. In \bibinfo{booktitle}{\emph{Proceedings of the 40th International Conference on Machine Learning}}.
\newblock


\bibitem[Gu and Dao(2023)]%
        {gu2023mamba}
\bibfield{author}{\bibinfo{person}{Albert Gu} {and} \bibinfo{person}{Tri Dao}.} \bibinfo{year}{2023}\natexlab{}.
\newblock \showarticletitle{Mamba: Linear-time sequence modeling with selective state spaces}.
\newblock \bibinfo{journal}{\emph{arXiv preprint arXiv:2312.00752}} (\bibinfo{year}{2023}).
\newblock


\bibitem[Hochreiter and Schmidhuber(1997)]%
        {hochreiter1997long}
\bibfield{author}{\bibinfo{person}{Sepp Hochreiter} {and} \bibinfo{person}{J{\"u}rgen Schmidhuber}.} \bibinfo{year}{1997}\natexlab{}.
\newblock \showarticletitle{Long short-term memory}.
\newblock \bibinfo{journal}{\emph{Neural computation}} \bibinfo{volume}{9}, \bibinfo{number}{8} (\bibinfo{year}{1997}), \bibinfo{pages}{1735--1780}.
\newblock


\bibitem[Huang et~al\mbox{.}(2019)]%
        {huang2019causal}
\bibfield{author}{\bibinfo{person}{Biwei Huang}, \bibinfo{person}{Kun Zhang}, \bibinfo{person}{Mingming Gong}, {and} \bibinfo{person}{Clark Glymour}.} \bibinfo{year}{2019}\natexlab{}.
\newblock \showarticletitle{Causal discovery and forecasting in nonstationary environments with state-space models}. In \bibinfo{booktitle}{\emph{International conference on machine learning}}. PMLR, \bibinfo{pages}{2901--2910}.
\newblock


\bibitem[Jasechko et~al\mbox{.}(2024)]%
        {jasechko2024rapid}
\bibfield{author}{\bibinfo{person}{Scott Jasechko}, \bibinfo{person}{Hansj{\"o}rg Seybold}, \bibinfo{person}{Debra Perrone}, \bibinfo{person}{Ying Fan}, \bibinfo{person}{Mohammad Shamsudduha}, \bibinfo{person}{Richard~G Taylor}, \bibinfo{person}{Othman Fallatah}, {and} \bibinfo{person}{James~W Kirchner}.} \bibinfo{year}{2024}\natexlab{}.
\newblock \showarticletitle{Rapid groundwater decline and some cases of recovery in aquifers globally}.
\newblock \bibinfo{journal}{\emph{Nature}} \bibinfo{volume}{625}, \bibinfo{number}{7996} (\bibinfo{year}{2024}), \bibinfo{pages}{715--721}.
\newblock


\bibitem[Kourakos et~al\mbox{.}(2019)]%
        {kourakos2019increasing}
\bibfield{author}{\bibinfo{person}{George Kourakos}, \bibinfo{person}{Helen~E Dahlke}, {and} \bibinfo{person}{Thomas Harter}.} \bibinfo{year}{2019}\natexlab{}.
\newblock \showarticletitle{Increasing groundwater availability and seasonal base flow through agricultural managed aquifer recharge in an irrigated basin}.
\newblock \bibinfo{journal}{\emph{Water Resources Research}} \bibinfo{volume}{55}, \bibinfo{number}{9} (\bibinfo{year}{2019}), \bibinfo{pages}{7464--7492}.
\newblock


\bibitem[Lahn et~al\mbox{.}(2024)]%
        {lahn2024combinatorial}
\bibfield{author}{\bibinfo{person}{Nathaniel Lahn}, \bibinfo{person}{Sharath Raghvendra}, {and} \bibinfo{person}{Kaiyi Zhang}.} \bibinfo{year}{2024}\natexlab{}.
\newblock \showarticletitle{A Combinatorial Algorithm for Approximating the Optimal Transport in the Parallel and MPC Settings}.
\newblock \bibinfo{journal}{\emph{Advances in Neural Information Processing Systems}}  \bibinfo{volume}{36} (\bibinfo{year}{2024}).
\newblock


\bibitem[Lam et~al\mbox{.}(2023)]%
        {lam2023learning}
\bibfield{author}{\bibinfo{person}{Remi Lam}, \bibinfo{person}{Alvaro Sanchez-Gonzalez}, \bibinfo{person}{Matthew Willson}, \bibinfo{person}{Peter Wirnsberger}, \bibinfo{person}{Meire Fortunato}, \bibinfo{person}{Ferran Alet}, \bibinfo{person}{Suman Ravuri}, \bibinfo{person}{Timo Ewalds}, \bibinfo{person}{Zach Eaton-Rosen}, \bibinfo{person}{Weihua Hu}, {et~al\mbox{.}}} \bibinfo{year}{2023}\natexlab{}.
\newblock \showarticletitle{Learning skillful medium-range global weather forecasting}.
\newblock \bibinfo{journal}{\emph{Science}} (\bibinfo{year}{2023}), \bibinfo{pages}{eadi2336}.
\newblock


\bibitem[Lan et~al\mbox{.}(2024a)]%
        {lan2024asynchronous}
\bibfield{author}{\bibinfo{person}{Guangchen Lan}, \bibinfo{person}{Dong-Jun Han}, \bibinfo{person}{Abolfazl Hashemi}, \bibinfo{person}{Vaneet Aggarwal}, {and} \bibinfo{person}{Christopher~G Brinton}.} \bibinfo{year}{2024}\natexlab{a}.
\newblock \showarticletitle{Asynchronous federated reinforcement learning with policy gradient updates: Algorithm design and convergence analysis}.
\newblock \bibinfo{journal}{\emph{arXiv preprint arXiv:2404.08003}} (\bibinfo{year}{2024}).
\newblock


\bibitem[Lan et~al\mbox{.}(2024b)]%
        {lan2024improved}
\bibfield{author}{\bibinfo{person}{Guangchen Lan}, \bibinfo{person}{Han Wang}, \bibinfo{person}{James Anderson}, \bibinfo{person}{Christopher Brinton}, {and} \bibinfo{person}{Vaneet Aggarwal}.} \bibinfo{year}{2024}\natexlab{b}.
\newblock \showarticletitle{Improved Communication Efficiency in Federated Natural Policy Gradient via ADMM-based Gradient Updates}.
\newblock \bibinfo{journal}{\emph{Advances in Neural Information Processing Systems}}  \bibinfo{volume}{36} (\bibinfo{year}{2024}).
\newblock


\bibitem[Levintal et~al\mbox{.}(2023)]%
        {levintal2023agricultural}
\bibfield{author}{\bibinfo{person}{Elad Levintal}, \bibinfo{person}{Maribeth~L Kniffin}, \bibinfo{person}{Yonatan Ganot}, \bibinfo{person}{Nisha Marwaha}, \bibinfo{person}{Nicholas~P Murphy}, {and} \bibinfo{person}{Helen~E Dahlke}.} \bibinfo{year}{2023}\natexlab{}.
\newblock \showarticletitle{Agricultural managed aquifer recharge (Ag-MAR)—a method for sustainable groundwater management: A review}.
\newblock \bibinfo{journal}{\emph{Critical Reviews in Environmental Science and Technology}} \bibinfo{volume}{53}, \bibinfo{number}{3} (\bibinfo{year}{2023}), \bibinfo{pages}{291--314}.
\newblock


\bibitem[Liegmann et~al\mbox{.}(2021)]%
        {liegmann2021real}
\bibfield{author}{\bibinfo{person}{Eyke Liegmann}, \bibinfo{person}{Petros Karamanakos}, {and} \bibinfo{person}{Ralph Kennel}.} \bibinfo{year}{2021}\natexlab{}.
\newblock \showarticletitle{Real-time implementation of long-horizon direct model predictive control on an embedded system}.
\newblock \bibinfo{journal}{\emph{IEEE Open Journal of Industry Applications}}  \bibinfo{volume}{3} (\bibinfo{year}{2021}), \bibinfo{pages}{1--12}.
\newblock


\bibitem[Liu et~al\mbox{.}(2023b)]%
        {liu2023financial}
\bibfield{author}{\bibinfo{person}{Shun Liu}, \bibinfo{person}{Kexin Wu}, \bibinfo{person}{Chufeng Jiang}, \bibinfo{person}{Bin Huang}, {and} \bibinfo{person}{Danqing Ma}.} \bibinfo{year}{2023}\natexlab{b}.
\newblock \showarticletitle{Financial time-series forecasting: Towards synergizing performance and interpretability within a hybrid machine learning approach}.
\newblock \bibinfo{journal}{\emph{arXiv preprint arXiv:2401.00534}} (\bibinfo{year}{2023}).
\newblock


\bibitem[Liu et~al\mbox{.}(2023a)]%
        {liu2023itransformer}
\bibfield{author}{\bibinfo{person}{Yong Liu}, \bibinfo{person}{Tengge Hu}, \bibinfo{person}{Haoran Zhang}, \bibinfo{person}{Haixu Wu}, \bibinfo{person}{Shiyu Wang}, \bibinfo{person}{Lintao Ma}, {and} \bibinfo{person}{Mingsheng Long}.} \bibinfo{year}{2023}\natexlab{a}.
\newblock \showarticletitle{iTransformer: Inverted Transformers Are Effective for Time Series Forecasting}.
\newblock \bibinfo{journal}{\emph{arXiv preprint arXiv:2310.06625}} (\bibinfo{year}{2023}).
\newblock


\bibitem[Liu et~al\mbox{.}(2022a)]%
        {liu2022non}
\bibfield{author}{\bibinfo{person}{Yong Liu}, \bibinfo{person}{Haixu Wu}, \bibinfo{person}{Jianmin Wang}, {and} \bibinfo{person}{Mingsheng Long}.} \bibinfo{year}{2022}\natexlab{a}.
\newblock \showarticletitle{Non-stationary Transformers: Exploring the Stationarity in Time Series Forecasting}.
\newblock  (\bibinfo{year}{2022}).
\newblock


\bibitem[Liu et~al\mbox{.}(2021)]%
        {liu2021video}
\bibfield{author}{\bibinfo{person}{Yilin Liu}, \bibinfo{person}{Shijia Zhang}, {and} \bibinfo{person}{Mahanth Gowda}.} \bibinfo{year}{2021}\natexlab{}.
\newblock \showarticletitle{When video meets inertial sensors: Zero-shot domain adaptation for finger motion analytics with inertial sensors}. In \bibinfo{booktitle}{\emph{Proceedings of the International Conference on Internet-of-Things Design and Implementation}}. \bibinfo{pages}{182--194}.
\newblock


\bibitem[Liu et~al\mbox{.}(2022b)]%
        {liu2022leveraging}
\bibfield{author}{\bibinfo{person}{Yilin Liu}, \bibinfo{person}{Shijia Zhang}, \bibinfo{person}{Mahanth Gowda}, {and} \bibinfo{person}{Srihari Nelakuditi}.} \bibinfo{year}{2022}\natexlab{b}.
\newblock \showarticletitle{Leveraging the properties of mmwave signals for 3d finger motion tracking for interactive iot applications}.
\newblock \bibinfo{journal}{\emph{Proceedings of the ACM on Measurement and Analysis of Computing Systems}} \bibinfo{volume}{6}, \bibinfo{number}{3} (\bibinfo{year}{2022}), \bibinfo{pages}{1--28}.
\newblock


\bibitem[Marwaha et~al\mbox{.}(2021)]%
        {marwaha2021identifying}
\bibfield{author}{\bibinfo{person}{Nisha Marwaha}, \bibinfo{person}{George Kourakos}, \bibinfo{person}{Elad Levintal}, {and} \bibinfo{person}{Helen~E Dahlke}.} \bibinfo{year}{2021}\natexlab{}.
\newblock \showarticletitle{Identifying agricultural managed aquifer recharge locations to benefit drinking water supply in rural communities}.
\newblock \bibinfo{journal}{\emph{Water Resources Research}} \bibinfo{volume}{57}, \bibinfo{number}{3} (\bibinfo{year}{2021}), \bibinfo{pages}{e2020WR028811}.
\newblock


\bibitem[Murphy(2022)]%
        {murphy2022examining}
\bibfield{author}{\bibinfo{person}{Nicholas~Paul Murphy}.} \bibinfo{year}{2022}\natexlab{}.
\newblock \bibinfo{booktitle}{\emph{Examining Nitrate Leaching Potential and Nitrogen Cycle Dynamics under Agricultural Managed Aquifer Recharge in the Central Valley of California}}.
\newblock \bibinfo{publisher}{University of California, Davis}.
\newblock


\bibitem[Nelson and Johnson(2020)]%
        {nelson2020model}
\bibfield{author}{\bibinfo{person}{James~R Nelson} {and} \bibinfo{person}{Nathan~G Johnson}.} \bibinfo{year}{2020}\natexlab{}.
\newblock \showarticletitle{Model predictive control of microgrids for real-time ancillary service market participation}.
\newblock \bibinfo{journal}{\emph{Applied Energy}}  \bibinfo{volume}{269} (\bibinfo{year}{2020}), \bibinfo{pages}{114963}.
\newblock


\bibitem[Nie et~al\mbox{.}(2022)]%
        {nie2022time}
\bibfield{author}{\bibinfo{person}{Yuqi Nie}, \bibinfo{person}{Nam~H Nguyen}, \bibinfo{person}{Phanwadee Sinthong}, {and} \bibinfo{person}{Jayant Kalagnanam}.} \bibinfo{year}{2022}\natexlab{}.
\newblock \showarticletitle{A Time Series is Worth 64 Words: Long-term Forecasting with Transformers}. In \bibinfo{booktitle}{\emph{The Eleventh International Conference on Learning Representations}}.
\newblock


\bibitem[Niswonger et~al\mbox{.}(2017)]%
        {niswonger2017managed}
\bibfield{author}{\bibinfo{person}{Richard~G Niswonger}, \bibinfo{person}{Eric~D Morway}, \bibinfo{person}{Enrique Triana}, {and} \bibinfo{person}{Justin~L Huntington}.} \bibinfo{year}{2017}\natexlab{}.
\newblock \showarticletitle{Managed aquifer recharge through off-season irrigation in agricultural regions}.
\newblock \bibinfo{journal}{\emph{Water Resources Research}} \bibinfo{volume}{53}, \bibinfo{number}{8} (\bibinfo{year}{2017}), \bibinfo{pages}{6970--6992}.
\newblock


\bibitem[O'Geen et~al\mbox{.}(2015)]%
        {o2015soil}
\bibfield{author}{\bibinfo{person}{AT O'Geen}, \bibinfo{person}{Matthew~BB Saal}, \bibinfo{person}{Helen~E Dahlke}, \bibinfo{person}{David~A Doll}, \bibinfo{person}{Rachel~B Elkins}, \bibinfo{person}{Allan Fulton}, \bibinfo{person}{Graham~E Fogg}, \bibinfo{person}{Thomas Harter}, \bibinfo{person}{Jan~W Hopmans}, \bibinfo{person}{Chuck Ingels}, {et~al\mbox{.}}} \bibinfo{year}{2015}\natexlab{}.
\newblock \showarticletitle{Soil suitability index identifies potential areas for groundwater banking on agricultural lands}.
\newblock \bibinfo{journal}{\emph{California Agriculture}} \bibinfo{volume}{69}, \bibinfo{number}{2} (\bibinfo{year}{2015}).
\newblock


\bibitem[{open-meteo}(2014)]%
        {openmeteo}
\bibfield{author}{\bibinfo{person}{{open-meteo}}.} \bibinfo{year}{2014}\natexlab{}.
\newblock \bibinfo{title}{{Free Weather API}}.
\newblock \bibinfo{howpublished}{\url{https://open-meteo.com/}}.
\newblock


\bibitem[Phatak et~al\mbox{.}(2023)]%
        {phatak2023computing}
\bibfield{author}{\bibinfo{person}{Abhijeet Phatak}, \bibinfo{person}{Sharath Raghvendra}, \bibinfo{person}{Chittaranjan Tripathy}, {and} \bibinfo{person}{Kaiyi Zhang}.} \bibinfo{year}{2023}\natexlab{}.
\newblock \showarticletitle{Computing all optimal partial transports}. In \bibinfo{booktitle}{\emph{International Conference on Learning Representations}}.
\newblock


\bibitem[Qiu et~al\mbox{.}(2019)]%
        {qiu2019nonlinear}
\bibfield{author}{\bibinfo{person}{Jiangxiao Qiu}, \bibinfo{person}{Samuel~C Zipper}, \bibinfo{person}{Melissa Motew}, \bibinfo{person}{Eric~G Booth}, \bibinfo{person}{Christopher~J Kucharik}, {and} \bibinfo{person}{Steven~P Loheide}.} \bibinfo{year}{2019}\natexlab{}.
\newblock \showarticletitle{Nonlinear groundwater influence on biophysical indicators of ecosystem services}.
\newblock \bibinfo{journal}{\emph{Nature Sustainability}} \bibinfo{volume}{2}, \bibinfo{number}{6} (\bibinfo{year}{2019}), \bibinfo{pages}{475--483}.
\newblock


\bibitem[Raghvendra et~al\mbox{.}(2024)]%
        {raghvendra2024new}
\bibfield{author}{\bibinfo{person}{Sharath Raghvendra}, \bibinfo{person}{Pouyan Shirzadian}, {and} \bibinfo{person}{Kaiyi Zhang}.} \bibinfo{year}{2024}\natexlab{}.
\newblock \showarticletitle{A New Robust Partial $ p $-Wasserstein-Based Metric for Comparing Distributions}.
\newblock \bibinfo{journal}{\emph{arXiv preprint arXiv:2405.03664}} (\bibinfo{year}{2024}).
\newblock


\bibitem[Rahmani and Frossard(2023)]%
        {rahmani2023castor}
\bibfield{author}{\bibinfo{person}{Abdellah Rahmani} {and} \bibinfo{person}{Pascal Frossard}.} \bibinfo{year}{2023}\natexlab{}.
\newblock \showarticletitle{Castor: Causal Temporal Regime Structure Learning}.
\newblock \bibinfo{journal}{\emph{arXiv preprint arXiv:2311.01412}} (\bibinfo{year}{2023}).
\newblock


\bibitem[Ren et~al\mbox{.}(2021)]%
        {ren2021rapt}
\bibfield{author}{\bibinfo{person}{Houxing Ren}, \bibinfo{person}{Jingyuan Wang}, \bibinfo{person}{Wayne~Xin Zhao}, {and} \bibinfo{person}{Ning Wu}.} \bibinfo{year}{2021}\natexlab{}.
\newblock \showarticletitle{Rapt: Pre-training of time-aware transformer for learning robust healthcare representation}. In \bibinfo{booktitle}{\emph{Proceedings of the 27th ACM SIGKDD conference on knowledge discovery \& data mining}}. \bibinfo{pages}{3503--3511}.
\newblock


\bibitem[Seth(2007)]%
        {seth2007granger}
\bibfield{author}{\bibinfo{person}{Anil Seth}.} \bibinfo{year}{2007}\natexlab{}.
\newblock \showarticletitle{Granger causality}.
\newblock \bibinfo{journal}{\emph{Scholarpedia}} \bibinfo{volume}{2}, \bibinfo{number}{7} (\bibinfo{year}{2007}), \bibinfo{pages}{1667}.
\newblock


\bibitem[Shen et~al\mbox{.}(2019)]%
        {shen2019deepapp}
\bibfield{author}{\bibinfo{person}{Zhihao Shen}, \bibinfo{person}{Kang Yang}, \bibinfo{person}{Wan Du}, \bibinfo{person}{Xi Zhao}, {and} \bibinfo{person}{Jianhua Zou}.} \bibinfo{year}{2019}\natexlab{}.
\newblock \showarticletitle{{DeepAPP: A Deep Reinforcement Learning Framework for Mobile Application Usage Prediction}}. In \bibinfo{booktitle}{\emph{ACM SenSys}}.
\newblock


\bibitem[Simnek et~al\mbox{.}(1999)]%
        {vsimuunek1999hydrus}
\bibfield{author}{\bibinfo{person}{J Simnek}, \bibinfo{person}{M Sejna}, {and} \bibinfo{person}{M~Th Van~Genuchten}.} \bibinfo{year}{1999}\natexlab{}.
\newblock \bibinfo{booktitle}{\emph{The HYDRUS-2D software package for simulating the two-dimensional movement of water, heat, and multiple solutes in variably-saturated media: Version 2.0}}.
\newblock \bibinfo{publisher}{US Salinity Laboratory, Agricultural Research Service}.
\newblock


\bibitem[Standen et~al\mbox{.}(2023)]%
        {standen2023integration}
\bibfield{author}{\bibinfo{person}{Kath Standen}, \bibinfo{person}{Lu{\'\i}s Costa}, \bibinfo{person}{Rui Hugman}, {and} \bibinfo{person}{Jos{\'e}~Paulo Monteiro}.} \bibinfo{year}{2023}\natexlab{}.
\newblock \showarticletitle{Integration of Managed Aquifer Recharge into the Water Supply System in the Algarve Region, Portugal}.
\newblock \bibinfo{journal}{\emph{Water}} \bibinfo{volume}{15}, \bibinfo{number}{12} (\bibinfo{year}{2023}), \bibinfo{pages}{2286}.
\newblock


\bibitem[USDA(1999)]%
        {usda1999united}
\bibfield{author}{\bibinfo{person}{NRCS USDA}.} \bibinfo{year}{1999}\natexlab{}.
\newblock \showarticletitle{United States department of agriculture}.
\newblock \bibinfo{journal}{\emph{Natural Resources Conservation Service. Plants Database. http://plants. usda. gov (accessed in 2000)}} (\bibinfo{year}{1999}).
\newblock


\bibitem[Vaswani et~al\mbox{.}(2017)]%
        {vaswani2017attention}
\bibfield{author}{\bibinfo{person}{Ashish Vaswani}, \bibinfo{person}{Noam Shazeer}, \bibinfo{person}{Niki Parmar}, \bibinfo{person}{Jakob Uszkoreit}, \bibinfo{person}{Llion Jones}, \bibinfo{person}{Aidan~N Gomez}, \bibinfo{person}{{\L}ukasz Kaiser}, {and} \bibinfo{person}{Illia Polosukhin}.} \bibinfo{year}{2017}\natexlab{}.
\newblock \showarticletitle{Attention is all you need}.
\newblock \bibinfo{journal}{\emph{Advances in neural information processing systems}}  \bibinfo{volume}{30} (\bibinfo{year}{2017}).
\newblock


\bibitem[Verma et~al\mbox{.}(2023)]%
        {verma2023climode}
\bibfield{author}{\bibinfo{person}{Yogesh Verma}, \bibinfo{person}{Markus Heinonen}, {and} \bibinfo{person}{Vikas Garg}.} \bibinfo{year}{2023}\natexlab{}.
\newblock \showarticletitle{ClimODE: Climate Forecasting With Physics-informed Neural ODEs}. In \bibinfo{booktitle}{\emph{The Twelfth International Conference on Learning Representations}}.
\newblock


\bibitem[Wang et~al\mbox{.}(2023a)]%
        {wang2023incremental}
\bibfield{author}{\bibinfo{person}{Dongjie Wang}, \bibinfo{person}{Zhengzhang Chen}, \bibinfo{person}{Yanjie Fu}, \bibinfo{person}{Yanchi Liu}, {and} \bibinfo{person}{Haifeng Chen}.} \bibinfo{year}{2023}\natexlab{a}.
\newblock \showarticletitle{Incremental causal graph learning for online root cause analysis}. In \bibinfo{booktitle}{\emph{Proceedings of the 29th ACM SIGKDD Conference on Knowledge Discovery and Data Mining}}. \bibinfo{pages}{2269--2278}.
\newblock


\bibitem[Wang et~al\mbox{.}(2023b)]%
        {wang2023interdependent}
\bibfield{author}{\bibinfo{person}{Dongjie Wang}, \bibinfo{person}{Zhengzhang Chen}, \bibinfo{person}{Jingchao Ni}, \bibinfo{person}{Liang Tong}, \bibinfo{person}{Zheng Wang}, \bibinfo{person}{Yanjie Fu}, {and} \bibinfo{person}{Haifeng Chen}.} \bibinfo{year}{2023}\natexlab{b}.
\newblock \showarticletitle{Interdependent Causal Networks for Root Cause Localization}. In \bibinfo{booktitle}{\emph{Proceedings of the 29th ACM SIGKDD Conference on Knowledge Discovery and Data Mining}}. \bibinfo{pages}{5051--5060}.
\newblock


\bibitem[Wu et~al\mbox{.}(2023)]%
        {wu2023timesnet}
\bibfield{author}{\bibinfo{person}{Haixu Wu}, \bibinfo{person}{Tengge Hu}, \bibinfo{person}{Yong Liu}, \bibinfo{person}{Hang Zhou}, \bibinfo{person}{Jianmin Wang}, {and} \bibinfo{person}{Mingsheng Long}.} \bibinfo{year}{2023}\natexlab{}.
\newblock \showarticletitle{TimesNet: Temporal 2D-Variation Modeling for General Time Series Analysis}. In \bibinfo{booktitle}{\emph{International Conference on Learning Representations(ICLR)}}.
\newblock


\bibitem[Xia et~al\mbox{.}(2023)]%
        {xia2023xcopy}
\bibfield{author}{\bibinfo{person}{Xianjin Xia}, \bibinfo{person}{Qianwu Chen}, \bibinfo{person}{Ningning Hou}, \bibinfo{person}{Yuanqing Zheng}, {and} \bibinfo{person}{Mo Li}.} \bibinfo{year}{2023}\natexlab{}.
\newblock \showarticletitle{XCopy: Boosting Weak Links for Reliable LoRa Communication}. In \bibinfo{booktitle}{\emph{Proceedings of the 29th Annual International Conference on Mobile Computing and Networking}}. \bibinfo{pages}{1--15}.
\newblock


\bibitem[Xu et~al\mbox{.}(2024)]%
        {xu2024cloudeval}
\bibfield{author}{\bibinfo{person}{Yifei Xu}, \bibinfo{person}{Yuning Chen}, \bibinfo{person}{Xumiao Zhang}, \bibinfo{person}{Xianshang Lin}, \bibinfo{person}{Pan Hu}, \bibinfo{person}{Yunfei Ma}, \bibinfo{person}{Songwu Lu}, \bibinfo{person}{Wan Du}, \bibinfo{person}{Zhuoqing Mao}, \bibinfo{person}{Ennan Zhai}, {et~al\mbox{.}}} \bibinfo{year}{2024}\natexlab{}.
\newblock \showarticletitle{CloudEval-YAML: A Practical Benchmark for Cloud Configuration Generation}.
\newblock \bibinfo{journal}{\emph{Proceedings of Machine Learning and Systems}}  \bibinfo{volume}{6} (\bibinfo{year}{2024}), \bibinfo{pages}{173--195}.
\newblock


\bibitem[Yang et~al\mbox{.}(2023)]%
        {yang2023link}
\bibfield{author}{\bibinfo{person}{Kang Yang}, \bibinfo{person}{Yuning Chen}, \bibinfo{person}{Xuanren Chen}, {and} \bibinfo{person}{Wan Du}.} \bibinfo{year}{2023}\natexlab{}.
\newblock \showarticletitle{Link quality modeling for lora networks in orchards}. In \bibinfo{booktitle}{\emph{Proceedings of the 22nd International Conference on Information Processing in Sensor Networks}}. \bibinfo{pages}{27--39}.
\newblock


\bibitem[Yang et~al\mbox{.}(2024a)]%
        {yang2024orchloc}
\bibfield{author}{\bibinfo{person}{Kang Yang}, \bibinfo{person}{Yuning Chen}, {and} \bibinfo{person}{Wan Du}.} \bibinfo{year}{2024}\natexlab{a}.
\newblock \showarticletitle{{OrchLoc: In-Orchard Localization via a Single LoRa Gateway and Generative Diffusion Model-based Fingerprinting}}. In \bibinfo{booktitle}{\emph{ACM MobiSys}}.
\newblock


\bibitem[Yang and Du(2022)]%
        {yang2022lldpc}
\bibfield{author}{\bibinfo{person}{Kang Yang} {and} \bibinfo{person}{Wan Du}.} \bibinfo{year}{2022}\natexlab{}.
\newblock \showarticletitle{{LLDPC: A Low-Density Parity-Check Coding Scheme for LoRa Networks}}. In \bibinfo{booktitle}{\emph{ACM SenSys}}.
\newblock


\bibitem[Yang and Du(2024)]%
        {yang2024low}
\bibfield{author}{\bibinfo{person}{Kang Yang} {and} \bibinfo{person}{Wan Du}.} \bibinfo{year}{2024}\natexlab{}.
\newblock \showarticletitle{{A Low-Density Parity-Check Coding Scheme for LoRa Networking}}.
\newblock \bibinfo{journal}{\emph{ACM Transactions on Sensor Networks}} (\bibinfo{year}{2024}).
\newblock


\bibitem[Yang et~al\mbox{.}(2024b)]%
        {yang2024rateless}
\bibfield{author}{\bibinfo{person}{Kang Yang}, \bibinfo{person}{Miaomiao Liu}, {and} \bibinfo{person}{Wan Du}.} \bibinfo{year}{2024}\natexlab{b}.
\newblock \showarticletitle{{RALoRa: Rateless-Enabled Link Adaptation for LoRa Networking}}.
\newblock \bibinfo{journal}{\emph{IEEE/ACM Transactions on Networking}} (\bibinfo{year}{2024}), \bibinfo{pages}{1--16}.
\newblock


\bibitem[Ye et~al\mbox{.}(2022)]%
        {ye2022learning}
\bibfield{author}{\bibinfo{person}{Junchen Ye}, \bibinfo{person}{Zihan Liu}, \bibinfo{person}{Bowen Du}, \bibinfo{person}{Leilei Sun}, \bibinfo{person}{Weimiao Li}, \bibinfo{person}{Yanjie Fu}, {and} \bibinfo{person}{Hui Xiong}.} \bibinfo{year}{2022}\natexlab{}.
\newblock \showarticletitle{Learning the evolutionary and multi-scale graph structure for multivariate time series forecasting}. In \bibinfo{booktitle}{\emph{Proceedings of the 28th ACM SIGKDD conference on knowledge discovery and data mining}}.
\newblock


\bibitem[Zeng et~al\mbox{.}(2023)]%
        {zeng2023transformers}
\bibfield{author}{\bibinfo{person}{Ailing Zeng}, \bibinfo{person}{Muxi Chen}, \bibinfo{person}{Lei Zhang}, {and} \bibinfo{person}{Qiang Xu}.} \bibinfo{year}{2023}\natexlab{}.
\newblock \showarticletitle{Are transformers effective for time series forecasting?}. In \bibinfo{booktitle}{\emph{Proceedings of the AAAI conference on artificial intelligence}}, Vol.~\bibinfo{volume}{37}. \bibinfo{pages}{11121--11128}.
\newblock


\bibitem[Zhang et~al\mbox{.}(2023)]%
        {zhang2023skilful}
\bibfield{author}{\bibinfo{person}{Yuchen Zhang}, \bibinfo{person}{Mingsheng Long}, \bibinfo{person}{Kaiyuan Chen}, \bibinfo{person}{Lanxiang Xing}, \bibinfo{person}{Ronghua Jin}, \bibinfo{person}{Michael~I Jordan}, {and} \bibinfo{person}{Jianmin Wang}.} \bibinfo{year}{2023}\natexlab{}.
\newblock \showarticletitle{Skilful nowcasting of extreme precipitation with NowcastNet}.
\newblock \bibinfo{journal}{\emph{Nature}} \bibinfo{volume}{619}, \bibinfo{number}{7970} (\bibinfo{year}{2023}), \bibinfo{pages}{526--532}.
\newblock


\end{thebibliography}

\appendix

\section{Sensor node deployment}\label{append:implement}

\begin{figure}[H]
	{\includegraphics[width=.48\textwidth]{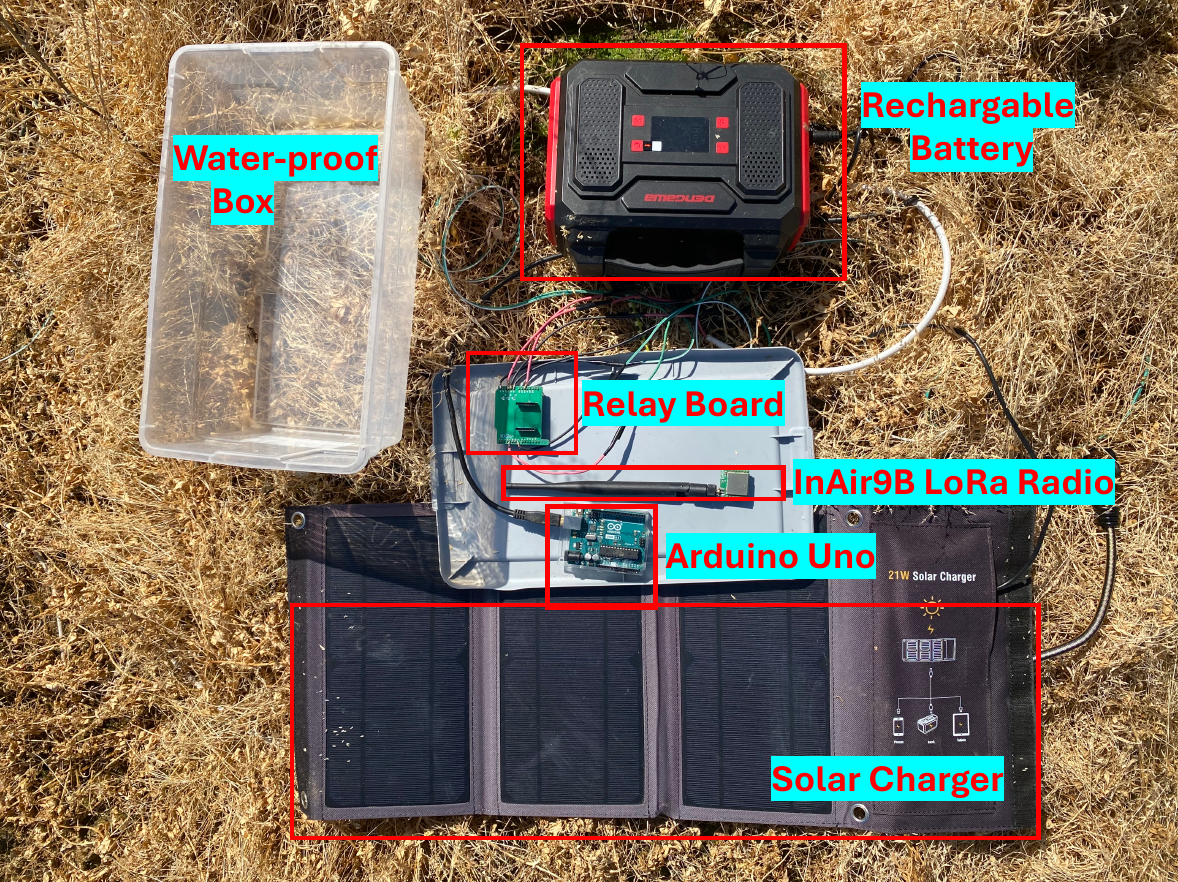}}
	\caption{The illustration of the solar-powered sensor node.}
	\label{fig_deploy_illus}
\end{figure}

The sensor node is powered by solar energy. As shown in Figure~\ref{fig_deploy_illus}, it consists of several key components. The Arduino Uno serves as the central controller, managing the operations of sensor readings and LoRa signal modulation. It is connected to KE-25 oxygen sensors and IRROMETER Watermark 200SS soil moisture sensors, which are deployed in the soil. The InAir9B LoRa radio is connected to the Arduino Uno via a relay board, which enables low-power long-range communication. The rechargeable battery along with the solar charger ensures minimum maintenance efforts. Key components are hosted in the waterproof box to protect them from damage and fast aging caused by environmental factors. The spreading factor (SF)~\cite{xia2023xcopy} of the LoRa transmission is 8.

\end{document}